\documentclass[a4paper,fleqn]{cas-dc}
\usepackage{natbib}
\usepackage{graphicx}
\usepackage[position=bottom]{subfig}
\usepackage{floatrow}
\usepackage{subfig}
\usepackage{float} 
\usepackage{booktabs}
\usepackage{acronym}
\usepackage[colorlinks]{hyperref}

\begin{document}
\let\WriteBookmarks\relax
\def\floatpagepagefraction{1}
\def\textpagefraction{.001}

\shorttitle{Precipitation Associated with Biparjoy }

\shortauthors{S. Verma et~al.}

\title [mode = title]{Estimation of the Area and Precipitation Associated with a Tropical Cyclone `Biparjoy' by using Image Processing}                      

%

\author[1,2]{Shikha Verma}[type=editor,
    orcid=0009-0003-5583-0431]
\cormark[1]
\ead{shikha.crest@gmail.com}

\author[1]{Kuldeep Srivastava}
\ead{kuldeep.srivastava@imd.gov.in}

\author[2]{Akhilesh Tiwari}
\ead{atiwari@iiita.ac.in}

\author[2]{Shekhar Verma}
\ead{sverma@iiita.ac.in}

\affiliation[1]{organization={India Meteorological Department},
    addressline={Ministry of Earth Sciences}, 
    state={New Delhi},
    country={India}}
    
\affiliation[2]{organization={Indian Institute of Information Technology, Allahabad},
    city={Prayagraj},
    state={Uttar Pradesh},
    country={India}}

\cortext[cor1]{Corresponding author}

\begin{abstract}
The rainfall associated with Topical Cyclone(TC) contributes a major amount to the annual rainfall in India. Due to the limited research on the quantitative precipitation associated with \ac{TC}, the prediction of the amount of precipitation and area that it may cover remains a challenge. This paper proposes an approach to estimate the accumulated precipitation and impact on affected area using Remote Sensing data. For this study, an instance of Extremely Severe Cyclonic Storm, ‘Biparjoy’ that formed over the Arabian Sea and hit India in 2023 is considered in which we have used the satellite images of \ac{IMERG}-Late Run of Global Precipitation Measurement (GPM). Image processing techniques were employed to identify and extract precipitation clusters linked to the cyclone. The results indicate that Biparjoy contributed a daily average rainfall of 53.14 mm/day across India and the Arabian Sea, with the Indian boundary receiving 11.59 mm/day, covering an extensive 411.76 thousand square kilometers. The localized intensity and variability observed in states like Gujarat, Rajasthan, Madhya Pradesh, and Uttar Pradesh highlight the need for tailored response measures, emphasizing the importance of further research to enhance predictive models and disaster readiness, crucial for building resilience against the diverse impacts of tropical cyclones.
 
\end{abstract}



\maketitle

\section{Introduction}
\ac{TCs} are one of the most destructive natural disasters that affect coastal regions worldwide \citep{ghosh2023cyclone}. The 5400 km coastline of India is affected by \ac{TC} of varying intensities. The North Indian Ocean (NIO) is a basin located between Africa and the Indian subcontinent and is one of the regions that experience \ac{TCs}. The \ac{NIO} basin is divided into two sub-basins, the Bay of Bengal (\ac{BoB}) and the Arabian Sea. The Bay of Bengal sub-basin is more active than the Arabian Sea, with the majority of the \ac{TCs} forming in this region. The \ac{NIO} basin experiences cyclones from April to December, with the peak season occurring from September to November. The cyclones in this region are classified by to the India Meteorological Department (\ac{IMD}) scale, which takes into account the maximum sustained wind speed near the center of the cyclone.

The impacts of \ac{TCs} on the \ac{NIO} region are significant, including damage to infrastructure, loss of life, and economic losses. The frequency and intensity of \ac{TCs} in the \ac{NIO} have been observed to be increasing in recent decades. This increase in \ac{TC} activity has been attributed to the rising sea surface temperatures, which provide favorable conditions for the formation and intensification of \ac{TCs} \citep{ali_vulnerability_1996, michaels_sea-surface_2006}. The \ac{NIO} region is also vulnerable to storm surges, which are caused by the combination of high winds and low atmospheric pressure during a \ac{TC}.  The impact of storm surges is often more severe than the impact of the strong winds associated with \ac{TCs}  \citep{jakobsen_comparison_2004, hubbert_real-time_1991, needham_review_2015}. Though the disaster caused by a \ac{TC} cannot be avoided, its impact can be greatly mitigated through a variety of management approaches, such as response, recovery, prevention/reduction, and preparedness \citep{hoque_tropical_2017}. However, this makes it essential to understand the patterns and behavior of \ac{TCs} in the \ac{NIO} basin to improve early warning systems.

Rainfall associated with \ac{TCs} has a significant impact on the environment and human activities in many regions of the world. These intense weather systems often cause severe flooding, landslides, and other hazards that can cause widespread damage to infrastructure, agriculture, and human settlements. Therefore, the study of rainfall patterns associated with \ac{TC} is crucial for understanding the characteristics and behavior of these storms and their potential impact on the affected areas. This may also allow meteorologists to prepare precipitation climatology by determining the average downpour over the globe. The long-term precipitation averages may be compared against short-term precipitation events to calculate deviation. These deviations or departures from climatology are an important aspect of interpreting the changing weather patterns. This will help to assess the severity of weather phenomena that are accompanied by severe precipitation occurrences such as \ac{TC}. 

It was observed that in 50 years (1970-2019), floods and tropical cyclones have been two major disasters causing mortality \citep{ray_assessment_2021}. One of the most significant impacts of tropical cyclones is the abundant rainfall produced by them \citep{marchok_validation_2007, rogers2009tropical}. While forecasts of \ac{TC} track and intensity have improved significantly, less attention has been paid to improve the estimate and forecast of the quantitative precipitation from \ac{TCs} \citep{marchok_validation_2007, biswas_case_2013, kidder_tropical_2005}. It has been observed that the precipitation associated with \ac{TCs} accounts for $\sim 5-6\%$ over the globe, $\sim 6-9\%$ of total precipitation over the Tropics, and about $5\%$ over the \ac{NIO} which includes both the Bay of Bengal (\ac{BoB}) and the Arabian Sea (AS) \citep{bhatla2020tropical, jiang_contribution_2010, lin2015tropical}.

Because of the dense population in peninsular India, the effects of \ac{TCs} rainfall are more severe over the \ac{NIO} \citep{singh_impact_2019}. The spatial distribution of rainfall \citep{osuri_customization_2012} and the time of occurrence of rainfall \citep{mohapatra_evaluation_2013} are sensitive to cyclone tracks. The \ac{IMD} and \ac{JTWC}, USA monitors the \ac{TCs} over \ac{NIO} to estimate the cyclone's path, intensity, landfall location, and time of landfall with the help of an optimal observational network, including satellite, radar, surface, and upper air observations \citep{mohapatra_best_2011}. 

The prediction of rainfall associated with a \ac{TC} is challenging because of \ac{TC}'s dynamic characteristics. Rainfall estimates based on manual observation, radar, \ac{NWP}, and other remote sensors present intriguing opportunities for improvement. However, there has been minimal prior research confirming the model QPF performance in \ac{TCs}. One of the key impediments to improving QPF in \ac{TCs} is the lack of a description of the distribution of rain in space and time \citep{tang_nowcasting_2018}. 

\citet{kidder_satellite_2000} estimated the rain rate in \ac{TC} with track forecasts using Tropical Rainfall Potential (TRaP) and calculated precipitation potential from Advanced Microwave Sounding Unit (\ac{AMSU}) data that appeared to be useful in forecasting the heavy precipitation associated with a landfalling \ac{TC}. \citet{lonfat_precipitation_2004} studied the precipitation distributions in \ac{TCs} from \ac{NASA}’s TRMM to understand the precipitation structure of tropical cyclones and the dynamics governing the rainfall structure. In an attempt by \cite{luitel_verification_2018}, the performance of \ac{NWP} was evaluated in forecasting rainfall associated with 15 U.S. landfalling \ac{TCs} during the 2007-2012 period. It was observed that there was a substantial error in the precipitation prediction associated with a \ac{TC}.  Hence, prior to the prediction, it is important to have a detailed study on quantitative analysis of rainfall observed during the lifetime of a \ac{TC}.  Understanding the quantitative aspects of rainfall, such as its intensity, volume, and distribution, meteorologists can better predict and assess the potential risks associated with these cyclones, enabling them to issue timely warnings and take appropriate mitigation measures. Additionally, it is crucial for climate studies and research. Climate change is known to influence the behavior of cyclones, including their rainfall characteristics. By analyzing and quantifying rainfall patterns over time, scientists can study the impact of climate change on tropical cyclones and assess any potential changes in their behavior, such as increased rainfall rates or shifts in the distribution of rainfall. By analyzing historical rainfall data and patterns, scientists can develop models that provide insights into the potential impact of rainfall associated with tropical cyclones, enabling authorities to issue timely warnings, plan evacuation strategies, and allocate necessary resources. Furthermore, it helps in assessing the hydrological risks associated with tropical cyclones, aiding in the design and implementation of appropriate infrastructure, drainage systems, and flood mitigation measures. 

One of the basic requirements of rainfall study associated with any weather phenomenon is to identify the area of proneness on the basis of vulnerability and hazard. Since areas affected by cyclonic rainfall can be influenced by a number of factors, such as environmental humidity, low-level vorticity, vertical wind shear, \ac{TC} latitude, and \ac{TC} intensity \citep{lin2015tropical}, it is important for any country to identify the prone locations in order to lower down the adverse impact of the cyclone. In this context, the government organizations of India such as the \ac{BMTPC} and \ac{NDMA} have listed vulnerable districts of different states of India based on cyclone-proneness. The \ac{IMD} has identified districts for which cyclone warnings must be issued. Taking it to further, \cite{mohapatra_classification_2012} have attempted to segregate the cyclone-prone districts of India into very high, high, moderate, and low categories of proneness. The categorization has been done on the basis of Probable Maximum Winds (PMW), Probable maximum storm surge (PMSS), and Probable maximum precipitation (PMP) for a day over any station in the district considering the other parameters such as enhanced rainfall in association with monsoon systems.

The manual observation of rainfall in the basin is not possible. Moreover, the weather radars installed at coasts have range limitations. Due to limited observations, it is difficult to accurately compute the actual rainfall that occurred over the basin during a cyclonic event. Fortunately, an earth-observing satellite \ac{GPM}, provides frequent estimates of precipitation at a global scale including rainfall occurred in the basin during a cyclonic period. The Integrated Multi-satellitE Retrievals for \ac{GPM} is designed to compensate for the limited sampling available from single \ac{LEO} -satellites by using as many \ac{LEO}-satellites as possible, and then augmenting with Geosynchronous-Earth-Orbit (\ac{GEO}) \ac{IR} estimates. The \ac{IMERG} system runs twice in near-real time providing “Late” multi-satellite product ~14 hr after observation time and is appropriate for daily and longer applications.

In this context, an attempt has been made to provide quantitative precipitation estimation using image processing techniques over \ac{IMERG} products and infer the complex relation between cyclones and associated precipitation. An instance of the Extremely Severe Cyclonic Storm, ‘Biparjoy’ is taken to explain the approach used in this study. The Biparjoy Cyclone occurred over the Arabian Sea and hit India in 2023. The daily average precipitation associated with Biparjoy over the Arabian Sea and India has been estimated along with the statistics of states affected by it.
 
The document begins by explaining the identified problem in Section \ref{problem_description}, shedding light on it's importance. Moving on to Section \ref{study_area}, it delves into the specific geographic focus of the study, while Section \ref{dataset} provides details about the data used. In Section \ref{methodology}, the step-by-step process employed to address the problem is detailed. The results are presented in Section \ref{result}, followed by an in-depth discussion in Section \ref{discussion}. Finally, the conclusion is presented in the Section \ref{conclusion}. 

\section{Problem Description} \label{problem_description}
Currently, there is a significant knowledge gap regarding \ac{TC}-induced rainfall patterns, hindering precise predictions and effective mitigation strategies. The lack of specific \ac{TC}-related rainfall data poses challenges in assessing the intensity and spatial distribution of precipitation within these cyclonic systems. To address this, a novel approach utilizing image processing techniques has been proposed to estimate \ac{TC}-associated rainfall and its spatial distribution. The segmentation of rainfall areas within the \ac{TC} system enables a quantitative assessment of rainfall extent and intensity. The significance of this research lies in improving our understanding of \ac{TC}-induced rainfall and enhancing the accuracy of rainfall predictions, aiding meteorologists, disaster management agencies, and policymakers in making informed decisions. Furthermore, this research focuses specifically on the Indian region, where tropical cyclones frequently make landfall and have the potential to cause significant damage. By developing a model that can accurately estimate the rainfall patterns associated with \ac{TCs} over India, this study contributes to the existing knowledge base and improves the overall understanding of cyclone dynamics in this region. This research can also serve as a foundation for further studies and advancements in \ac{TC} rainfall estimation, ultimately leading to more accurate predictions and better preparedness for cyclone-related rainfall events. However, the reliance on satellite imagery introduces limitations such as potential cloud cover, image resolution, and real-time data availability, which can affect the accuracy and timeliness of rainfall estimates. Addressing these limitations is crucial for robust and accurate \ac{TC} rainfall estimations.

\section{Study Area and Dataset}
\begin{figure}[bp]
	\centering
		\includegraphics[scale=0.30]{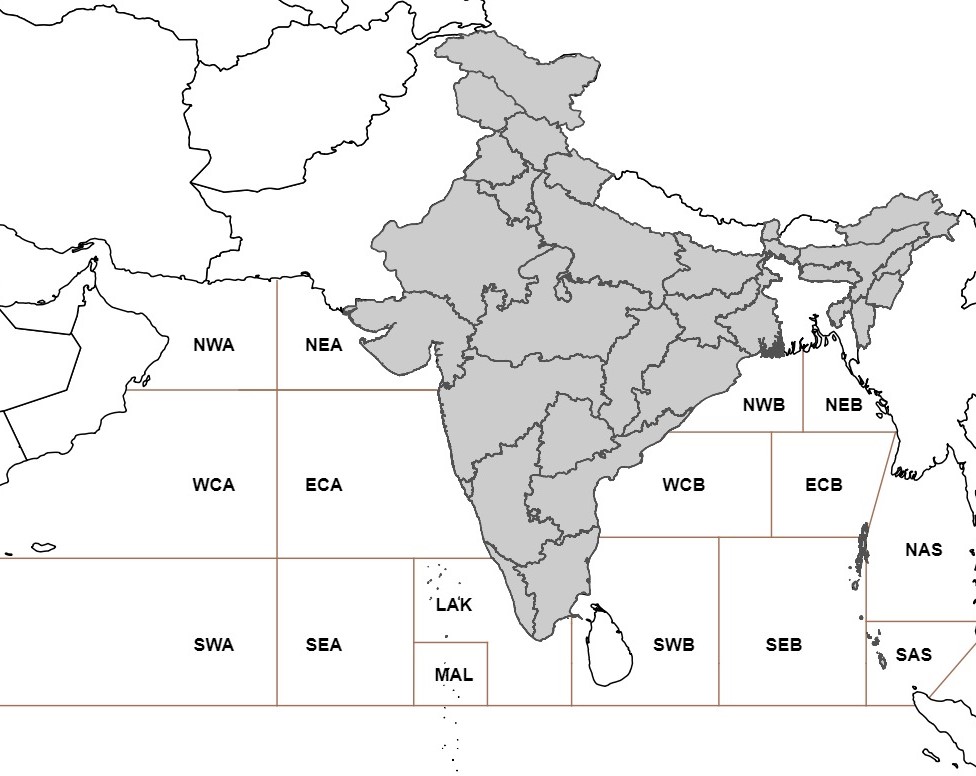}
	  \caption{Study area}
        \label{fig1}
\end{figure}

\subsection{Study Area} \label{study_area}
The study area chosen for this research is the Arabian Sea (AS) of the \ac{NIO} basin as shown in shaded region of Figure \hyperref[fig1]{\ref*{fig1}}. 
The AS is divided into 6 zones namely NWA, NEA, WCA, ECA, SWA, and SEA. The region is known to be prone to tropical cyclones, particularly during the pre-monsoon and post-monsoon seasons, and therefore, it is important to understand the behavior of these cyclones to better prepare and mitigate their impacts. The cyclogenesis of tropical cyclone Biparjoy of Extremely Severe Cyclonic Storm (\ac{ESCS}) category was observed in AS on 06-06-2023, which was the first \ac{CS} over the Arabian Sea in the year 2023. As per the report of \ac{IMD}, it crossed Saurashtra and Kutch coast close to Jakhau Port near latitude 23.28°N and longitude 68.56°E with \ac{MSW} of 65 knots gusting to 75 knots on 15-06-2023. Finally, it weakened into a depression (\ac{D}) over South Rajasthan and adjoining North Gujarat on 18-06-2023 and into a WML over central parts of Northeast Rajasthan and neighbourhood on 19-06-2023 at 03 UTC.

\subsection{Dataset} \label{dataset}

The GPM\_3IMERGHHL (\ac{GPM} \ac{IMERG} Late Precipitation L3 Half Hourly) dataset has been selected as the primary data set for this study. This choice is motivated by the dataset's capability to provide comprehensive and detailed information on precipitation dynamics. Utilizing advanced algorithms such as GPROF2017, the dataset offers half-hourly precipitation estimates with a spatial resolution of 0.1° x 0.1°. Notably, the inclusion of "late" observations, occurring around 14 hours after the initial observation time, adds temporal depth to the dataset. The meticulous processing, including gridding, intercalibration to the \ac{GPM} Combined Ku Radar-Radiometer Algorithm (CORRA) product, and bias correction, enhances the dataset's accuracy and reliability, making it an ideal choice for the study's comprehensive analysis of precipitation patterns \citep{NASA_GES_DISC}.

In adherence to the WMO standards, this study considers the Date of Acquisition (DOA) for 24-hour precipitation from 07-06-2023 to 20-06-2023 recorded at 03 UTC to ensure a consistent and uniform observation protocol. The detail of the dataset along with Date\_id (D\_ID) is listed in Table \hyperref[tbl1]{\ref*{tbl1}} and will be used as a nomenclature in the further sections.

\begin{table}[tp]
\caption{Details of the dataset used}\label{tbl1}
\centering
\begin{tabular}{@{}lll@{}}
\toprule
SN & \ac{DOA} & D\_ID \\
\midrule
1	&	07-06-2023	&	D1	\\
2	&	08-06-2023	&	D2	\\
3	&	09-06-2023	&	D3	\\
4	&	10-06-2023	&	D4	\\
5	&	11-06-2023	&	D5	\\
6	&	12-06-2023	&	D6	\\
7	&	13-06-2023	&	D7	\\
8	&	14-06-2023	&	D8	\\
9	&	15-06-2023	&	D9	\\
10	&	16-06-2023	&	D10	\\
11	&	17-06-2023	&	D11	\\
12	&	18-06-2023	&	D12	\\
13	&	19-06-2023	&	D13	\\
14	&	20-06-2023	&	D14	\\
\bottomrule
\end{tabular}
\end{table}

\section{Methodology} \label{methodology}
The proposed method uses satellite imagery to analyze rainfall and determine the extent of the Biparjoy cyclone's coverage. By processing the satellite data using image processing techniques, we can identify specific rainfall patterns and understand the impact of the cyclone more accurately. This approach overcomes the limitations of traditional methods and provides a clearer picture of the cyclone's effects, even in areas with limited ground-based observations. Hence, it offers a more comprehensive understanding of the cyclone's impact and enables improved decision-making and response strategies.

\begin{figure}[tp]
	\centering
		\includegraphics[width=60mm]{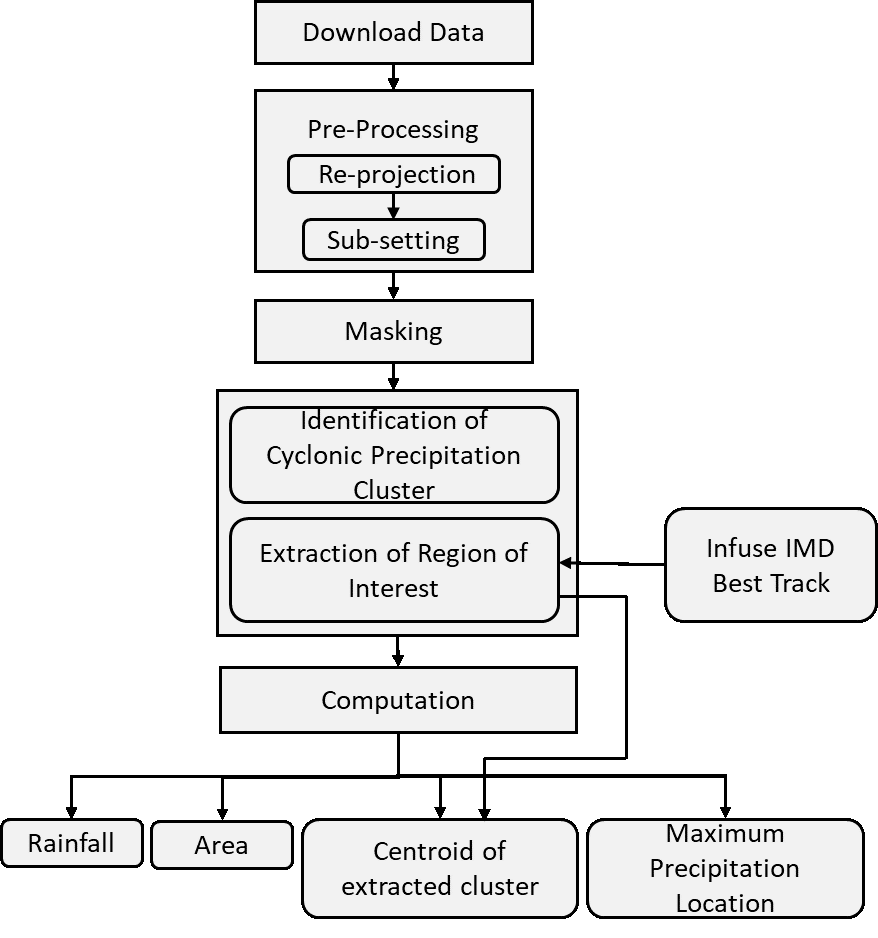}
	  \caption{Workflow of the proposed approach}
        \label{fig2}
\end{figure}

The step-wise workflow of the proposed approach is shown in Figure \hyperref[fig2]{\ref*{fig2}}. It shows that the data is downloaded which is pre-processed for further processing. Then, the mask of the precipitation blob has been generated from which the major blob was extracted based on the connected component. The precipitation from the extracted cluster has been computed for the study site. Based on the spatial resolution of \ac{IMERG} data of rainfall, the total area affected by rainfall associated with cyclone ‘Biparjoy’ in India has been computed. A comprehensive explanation of each step is given below.

\subsection{Data download}	
The \ac{IMERG} product GPM\_3IMERHHL v06 is downloaded from the data server of \ac{NASA}. The details of the dataset are given in section \ref{dataset}. IMERG (Integrated Multi-satellite Retrievals for GPM) is a unified satellite precipitation product that provides estimated surface precipitation over most of the globe. It is a product of Global Precipitation Measurement (GPM), which is an international satellite mission jointly launched by \ac{NASA} and \ac{JAXA} to provide unified precipitation measurements of rain and snow every three hours.

\subsection{Pre-processing}
In this section, the pre-processing steps for the downloaded data are explained. The pre-processing steps ensure that the data is in the appropriate format and ready for further analysis. These steps are important in any analysis involving geospatial data, as they can significantly impact the accuracy and reliability of the results.

The first step in pre-processing is re-projection, which involves warping the images from their original projection space to a selected projection space. This process requires the calculation of a geometric transformation model, which accounts for the relative geometry of the two projections and local scale and resolution changes in the images. The amount of distortion incurred during re-projection depends on these factors. In this study, the projection of downloaded raster data is converted from sinusoidal to geographic latitude-longitude for further processing \citep{steinwand1995map}. 

\begin{figure}
    \centering
    \subfloat[]{{\includegraphics[width=2.5cm]{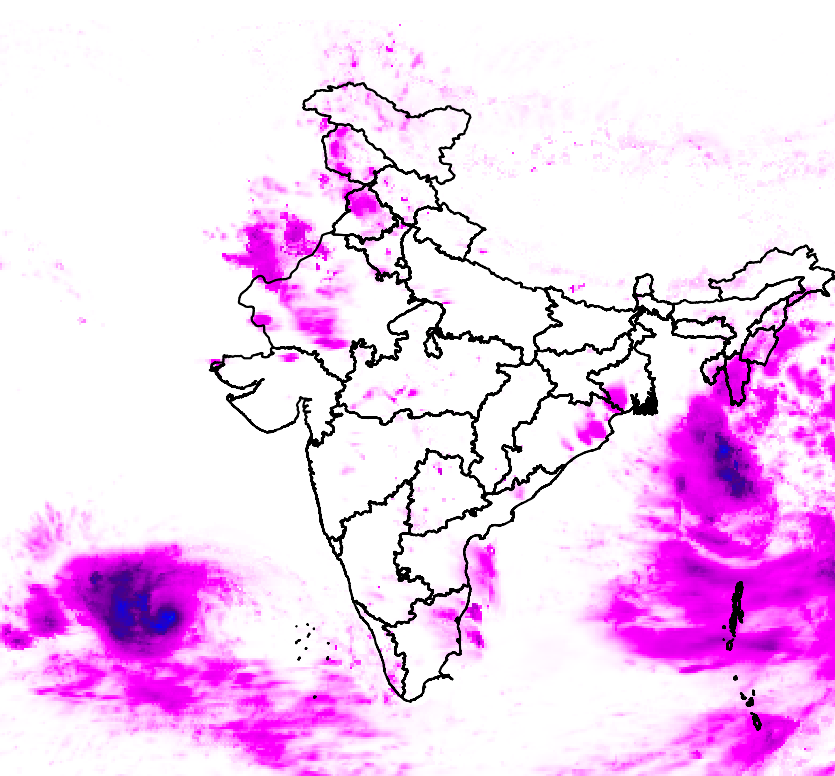} }} 
    \subfloat[]{{\includegraphics[width=2.5cm]{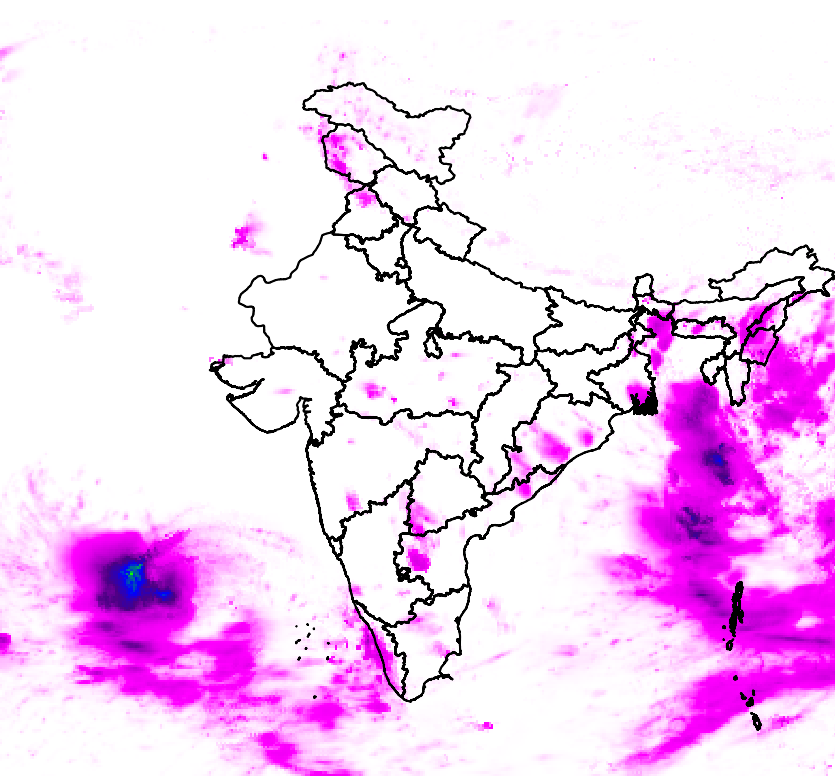} }}     
    \subfloat[]{{\includegraphics[width=2.5cm]{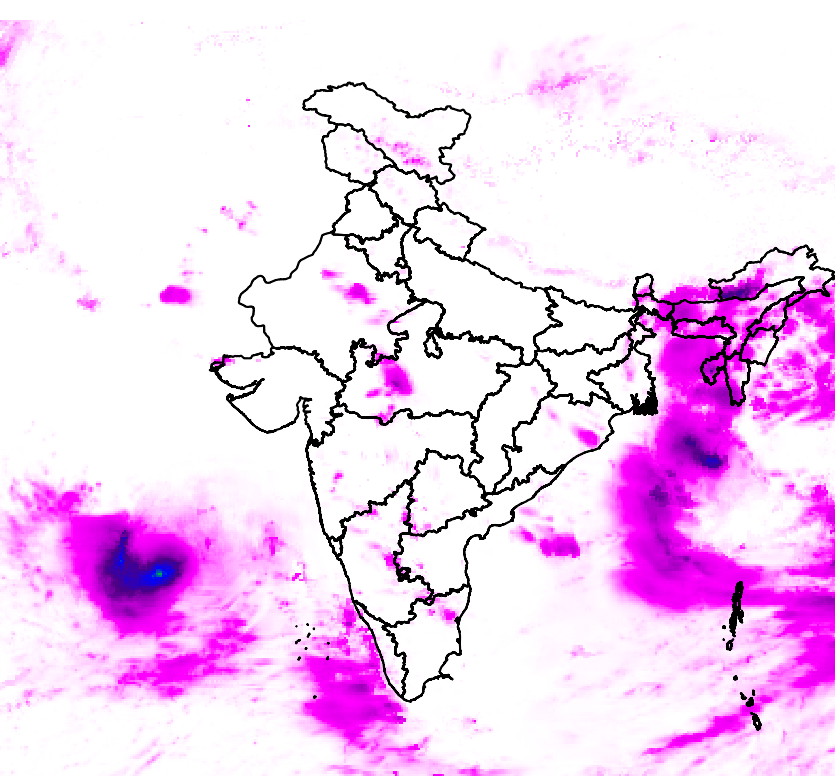} }} 
    
    \subfloat[]{{\includegraphics[width=2.5cm]{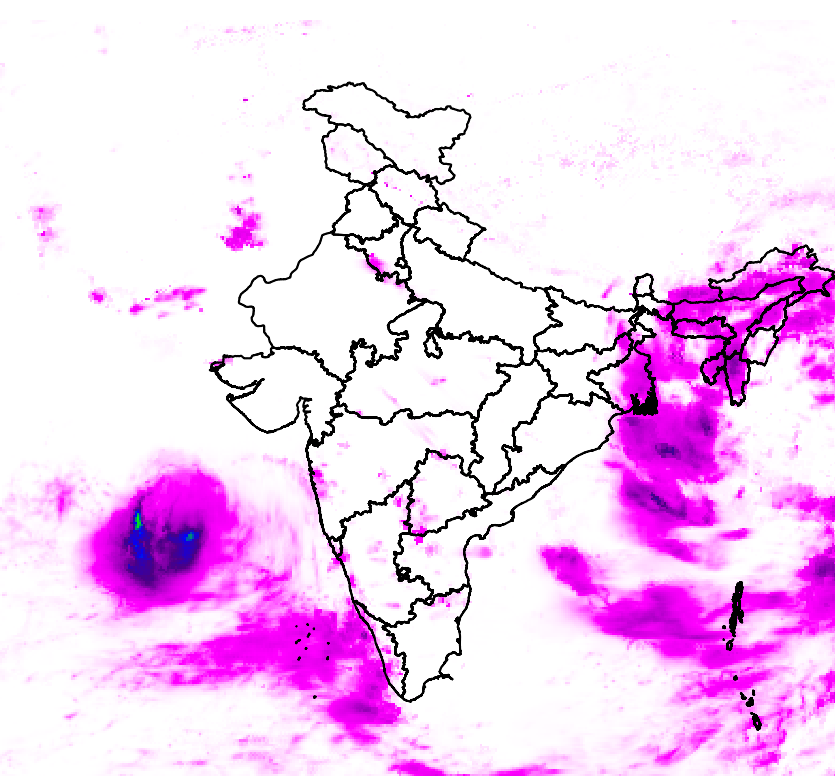} }} 
    \subfloat[]{{\includegraphics[width=2.5cm]{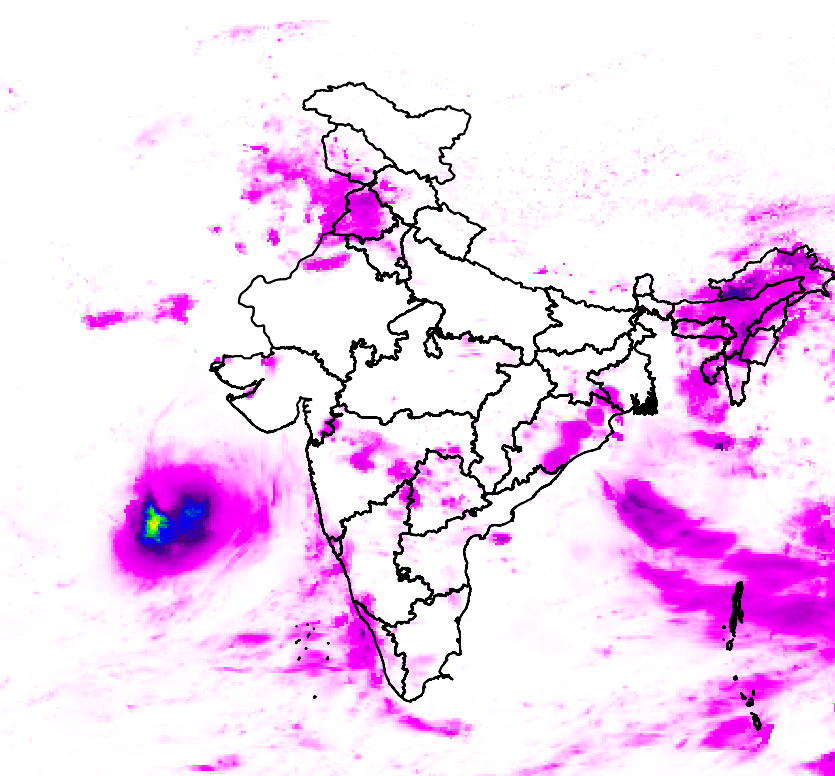} }}     
    \subfloat[]{{\includegraphics[width=2.5cm]{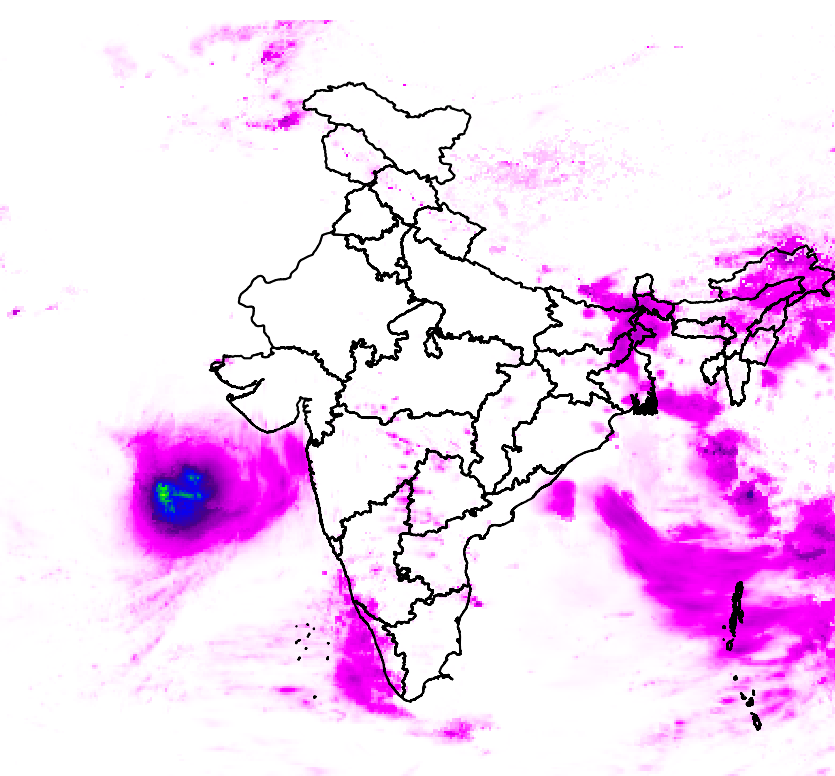} }} 
    
    \subfloat[]{{\includegraphics[width=2.5cm]{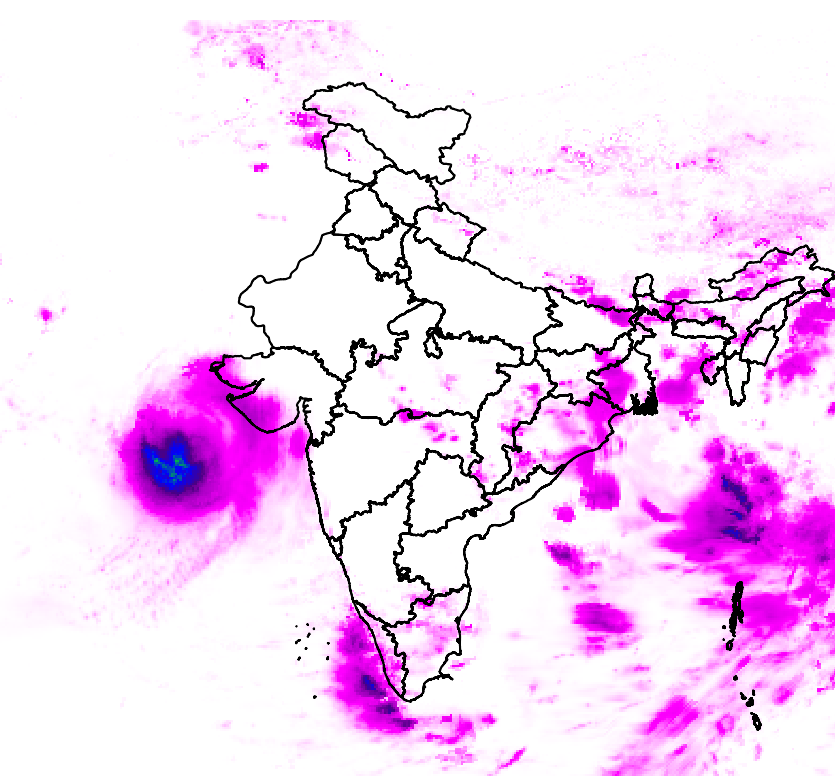} }} 
    \subfloat[]{{\includegraphics[width=2.5cm]{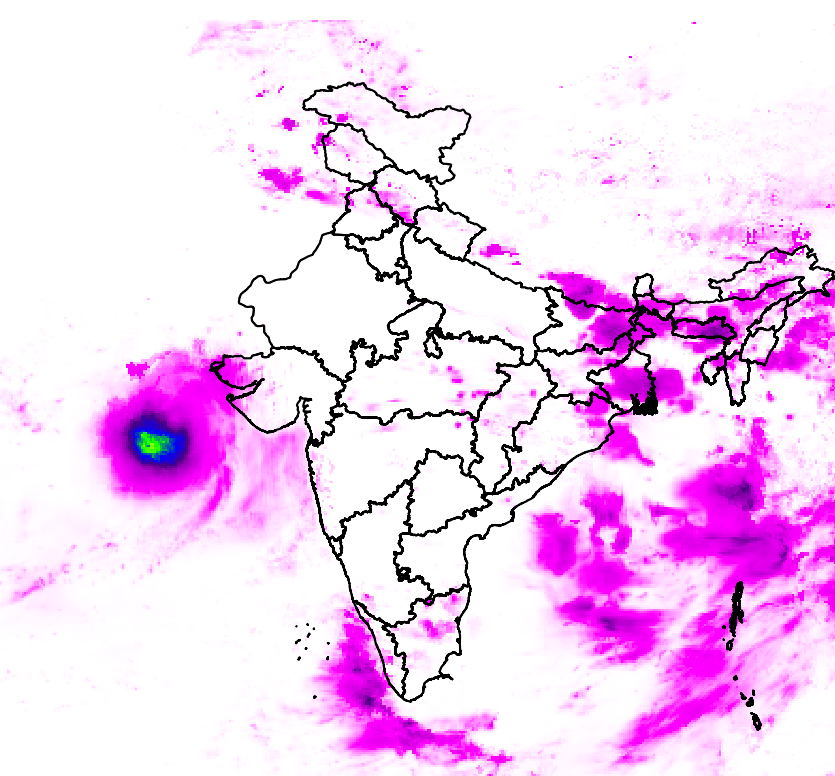} }}     
    \subfloat[]{{\includegraphics[width=2.5cm]{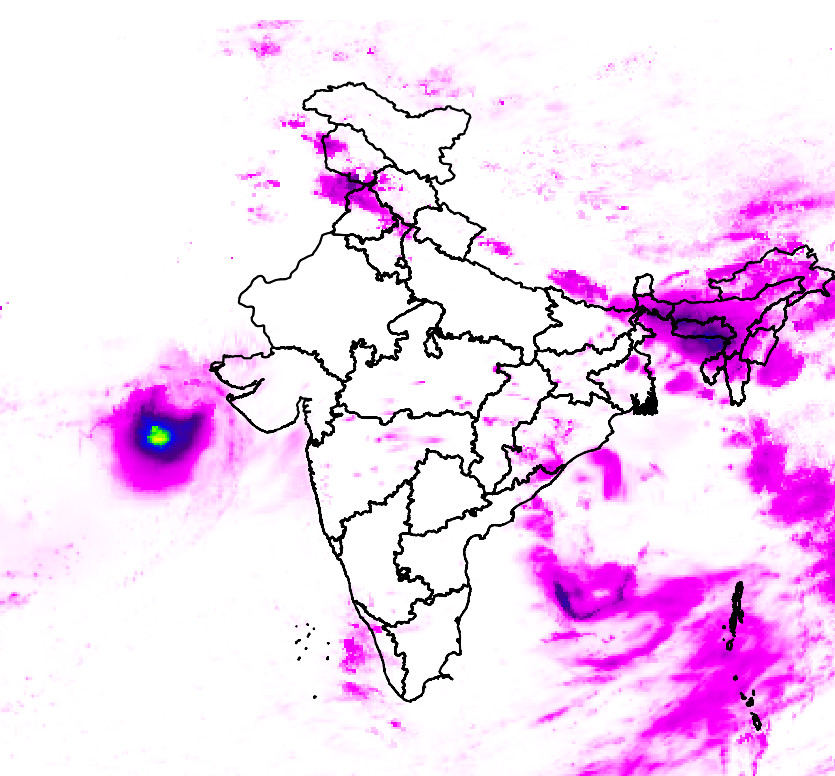} }} 
    
    \subfloat[]{{\includegraphics[width=2.5cm]{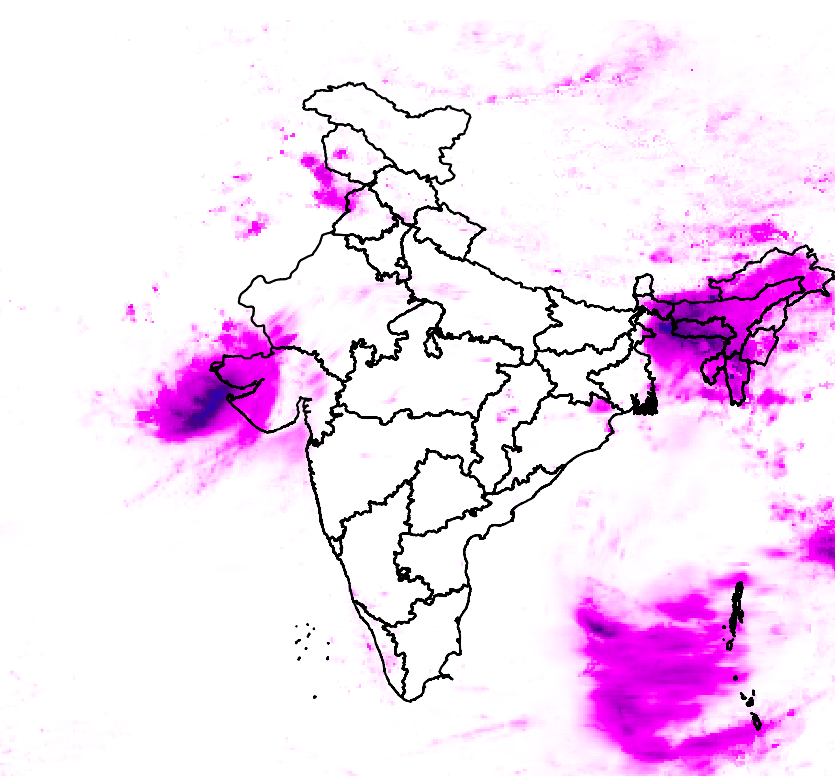} }} 
    \subfloat[]{{\includegraphics[width=2.5cm]{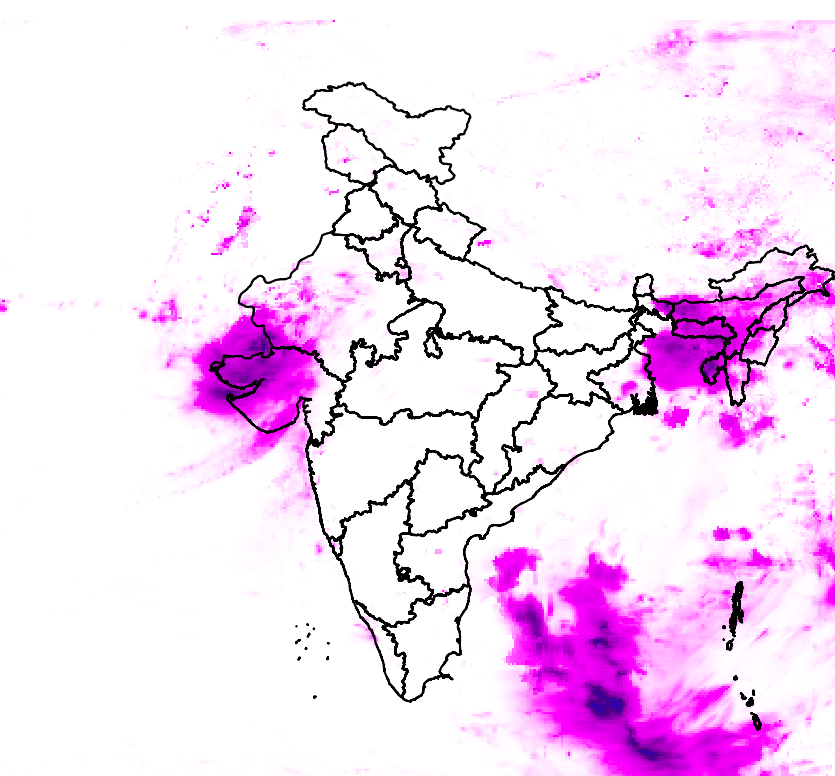} }}     
    \subfloat[]{{\includegraphics[width=2.5cm]{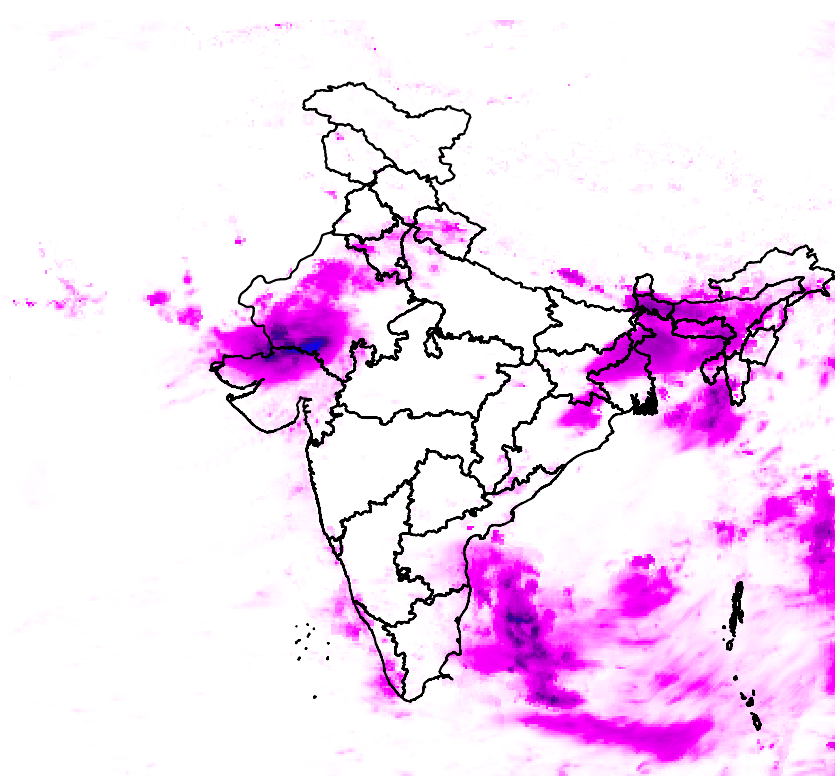} }} 
    
    \subfloat[]{{\includegraphics[width=2.5cm]{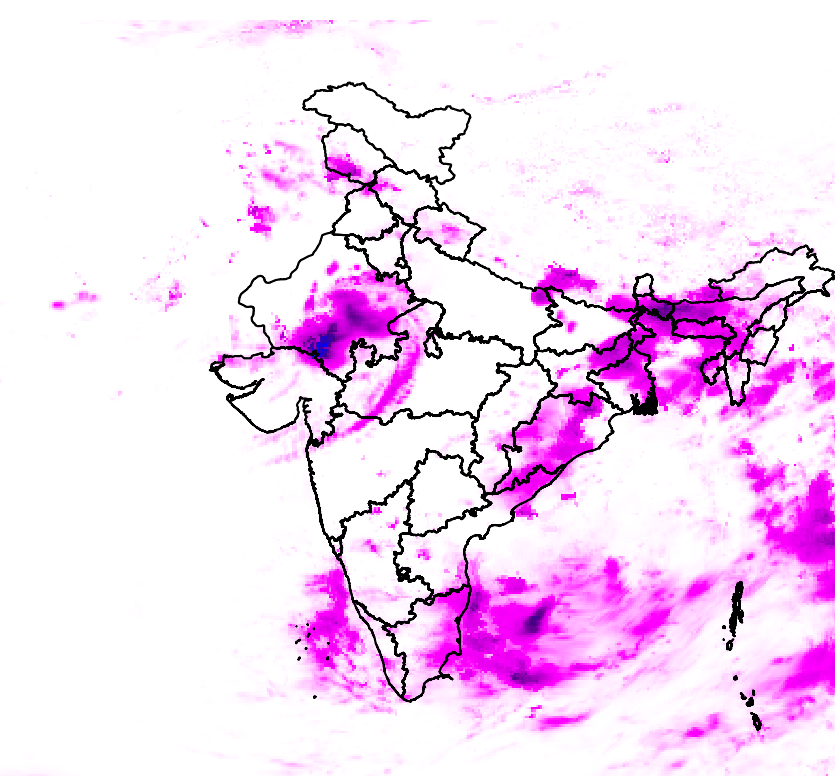} }} 
    \subfloat[]{{\includegraphics[width=2.5cm]{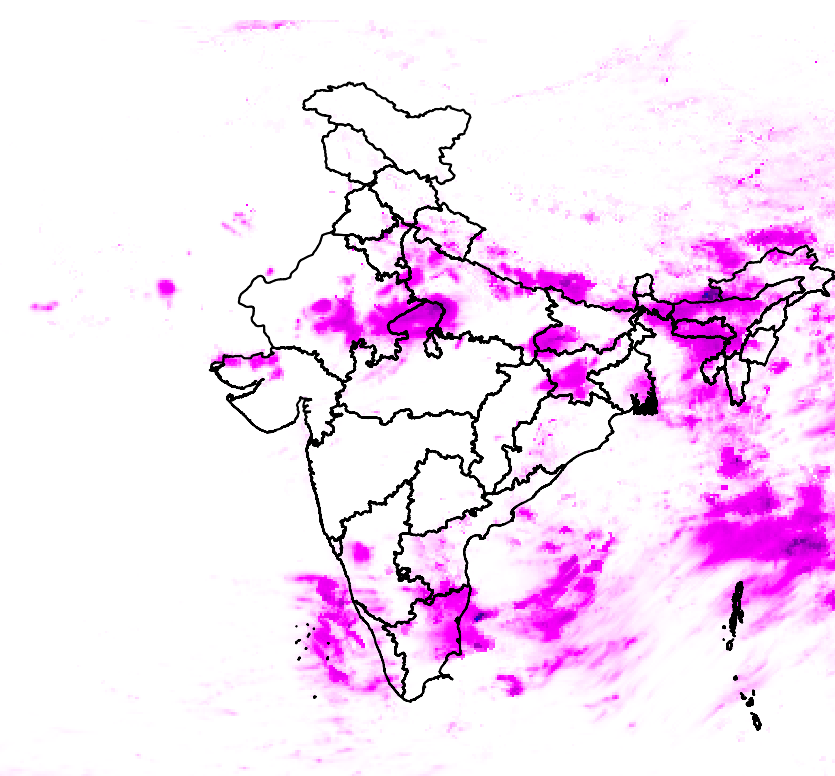} }} 
    
    \caption{Pre-processed images for the D\_ID (a)-(n): D1-D14}
    \label{fig3}
\end{figure}

The next step in pre-processing is spatial subsetting, which involves extracting the region of interest (ROI). This process returns a new object with only features that are spatially related to the \ac{ROI} \citep{ayoobi_evaluating_2017}. This step is important because it reduces the data size and processing time, making it easier to analyze the data. Figure \hyperref[fig3]{\ref*{fig3}} shows the images for D1 to D14 after re-projection and spatial subsetting. 

\subsection{Masking}
\begin{figure}
    \centering
    \subfloat[]{{\includegraphics[width=2.5cm]{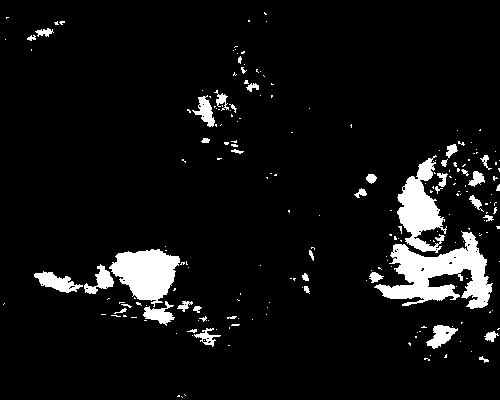} }} 
    \subfloat[]{{\includegraphics[width=2.5cm]{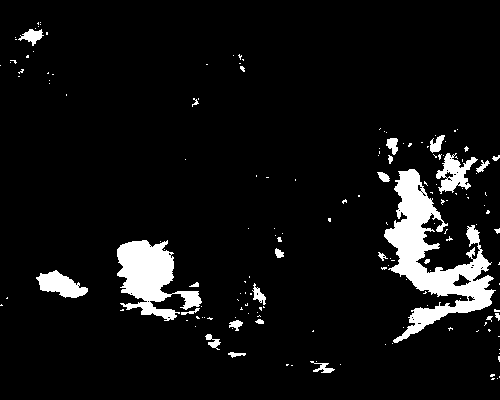} }}
    \subfloat[]{{\includegraphics[width=2.5cm]{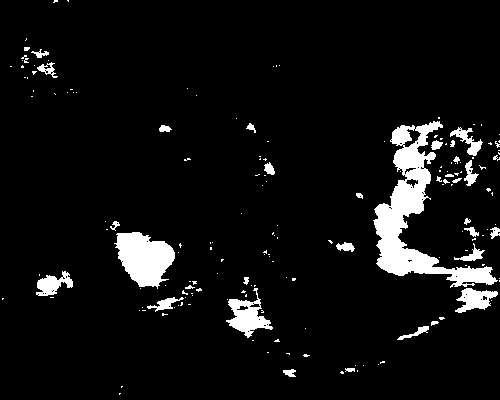} }}
    
    \subfloat[]{{\includegraphics[width=2.5cm]{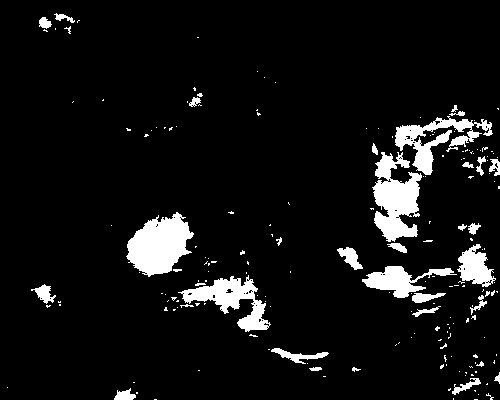} }}    
    \subfloat[]{{\includegraphics[width=2.5cm]{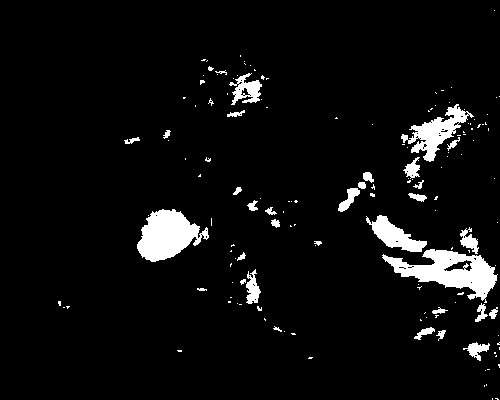} }}
    \subfloat[]{{\includegraphics[width=2.5cm]{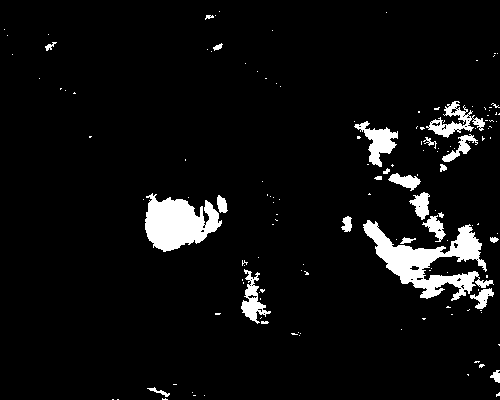} }}
    
    \subfloat[]{{\includegraphics[width=2.5cm]{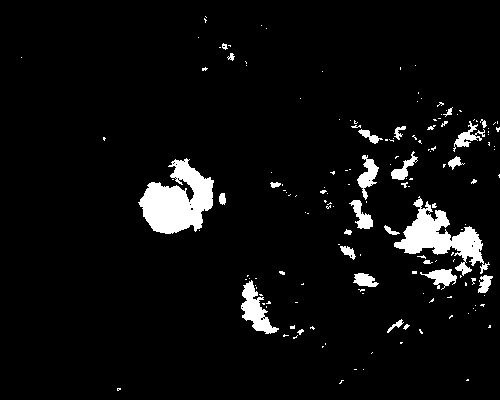} }}    
    \subfloat[]{{\includegraphics[width=2.5cm]{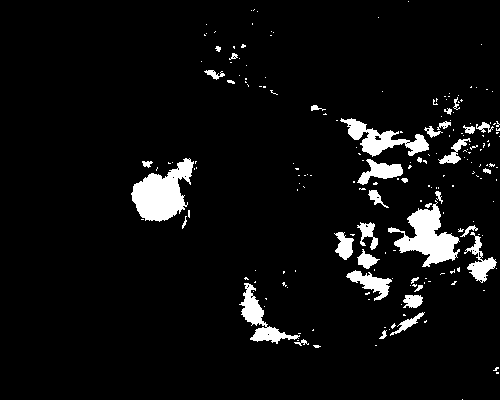} }}
    \subfloat[]{{\includegraphics[width=2.5cm]{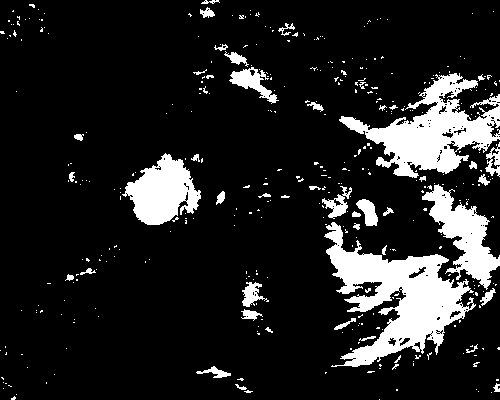} }}  

    \subfloat[]{{\includegraphics[width=2.5cm]{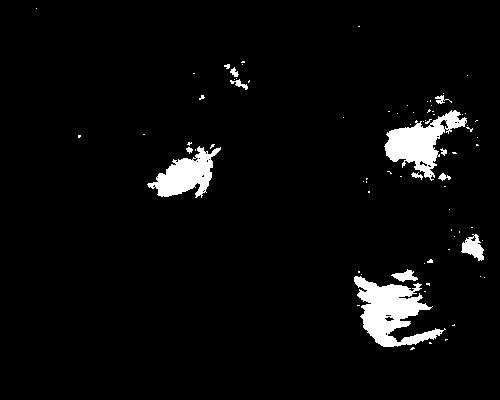} }}    
    \subfloat[]{{\includegraphics[width=2.5cm]{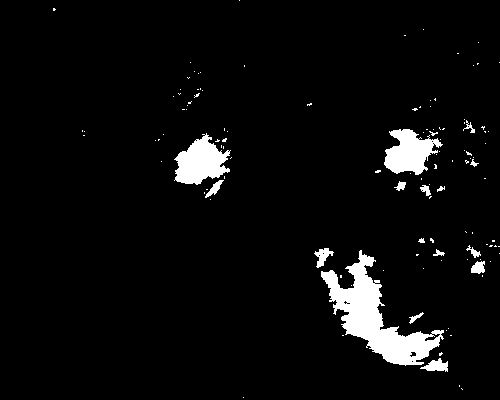} }}
    \subfloat[]{{\includegraphics[width=2.5cm]{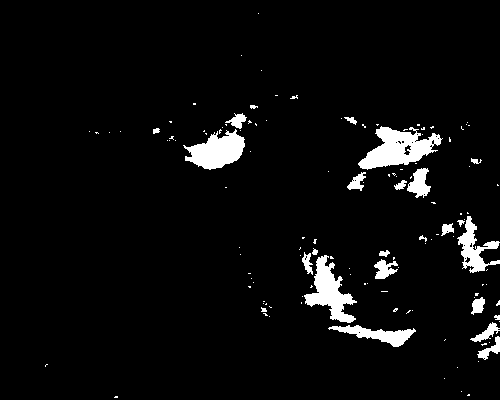} }}  
    
    \subfloat[]{{\includegraphics[width=2.5cm]{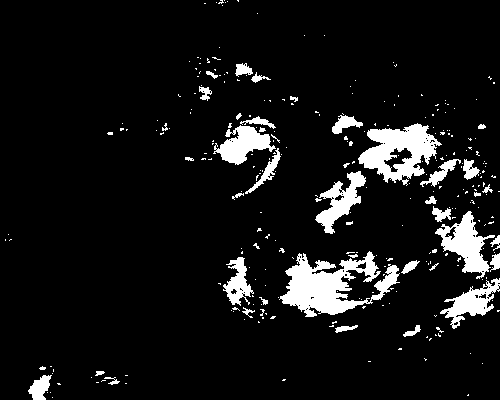} }}
    \subfloat[]{{\includegraphics[width=2.5cm]{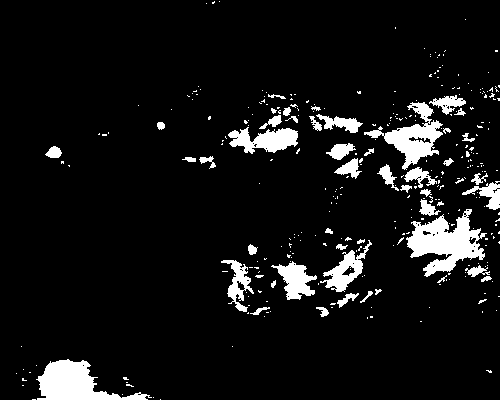} }} 
    
    \caption{Cloud mask for D\_ID (a)-(n): D1-D14}%
    \label{fig4}
\end{figure}

 After the pre-processing step, a two-step thresholding approach is employed to filter out irrelevant information and highlight the crucial features in the satellite imagery. Initially, pixels with values exceeding 0.9 mm are masked out, focusing on regions where precipitation is significant. Subsequently, a finer threshold of 0.3 mm is employed, emphasizing the areas with a more pronounced cyclonic structure. This sequential approach ensures that the masking strategy evolves from a broader emphasis on significant precipitation to a more nuanced focus, particularly when the cyclone's eye becomes prominent. The outcome of this process is a binary mask image, which contains pixels with either a value of 0 or 1, representing the areas that meet the specified condition and those that do not. To visually represent the effectiveness of this preprocessing step, the masked images for the D1 to D14 are shown in Figure \hyperref[fig4]{\ref*{fig4}}.

\subsection{Identification and Extraction of Cyclonic Precipitation Cluster}
In this step of the study, the focus is on labeling the results obtained from the previous step based on the area they cover. The objective is to identify the precipitation clusters associated with the Biparjoy cyclone, which can then be further processed. The employed methodology involves leveraging Connected Component Analysis (\ac{CCA}). This technique is utilized to effectively identify and label connected regions within a binary image, enabling a detailed examination of the precipitation patterns associated with the cyclone. It involves finding the objects in a binary image by creating a labeled image where pixels with identical integer values belong to the same class of object. The algorithm works by scanning the binary image pixel by pixel and checking whether each pixel is connected to any of its neighbors. A pixel is considered connected to its neighbors if they have the same value, and they are adjacent to each other either horizontally or vertically. When a group of connected pixels is identified, they are assigned a unique label, and the process is repeated until all pixels in the image have been labeled. The result of the \ac{CCA} algorithm is a labeled image, where each cluster of connected pixels is assigned a unique label. This labeled image can then be used for further analysis and processing. Mathematically, the \ac{CCA} algorithm can be represented by the following equation: 
\[ L = bwlabel(BW) \] 
where L is the labeled image, BW is the binary image, and bwlabel is the function that performs the \ac{CCA} operation. The function bwlabel assigns a unique integer label to each connected cluster of pixels in the binary image BW.

In the present study, the binary image represents the precipitation clusters associated with the cyclone. The False value in the binary image represents background pixels, while the True value represents object pixels. 

\begin{figure}
	\centering
		\includegraphics[width=70mm]{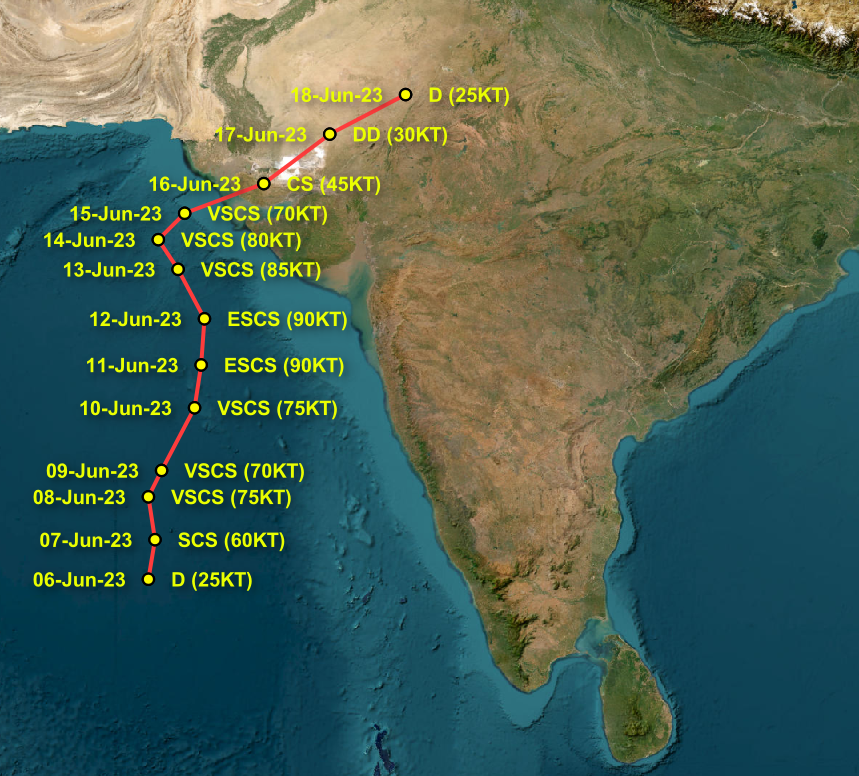}
	  \caption{Best Track of Biprajoy Cyclone}
        \label{fig5}
\end{figure}

Moreover, to ascertain the precise association of identified precipitation clusters with the Biparjoy cyclone and to mitigate any potential ambiguity arising from other weather events in the region, we are incorporating the best track information of 03 UTC provided by the \ac{IMD}, as illustrated in Figure \hyperref[fig5]{\ref*{fig5}}. This meticulous consideration of the Biparjoy cyclone's best track is essential for the accurate labeling of the identified precipitation clusters, laying a crucial foundation for the subsequent in-depth analysis of the associated rainfall patterns.

\begin{figure}[ht]
    \centering
    \subfloat[]{{\includegraphics[width=2.5cm]{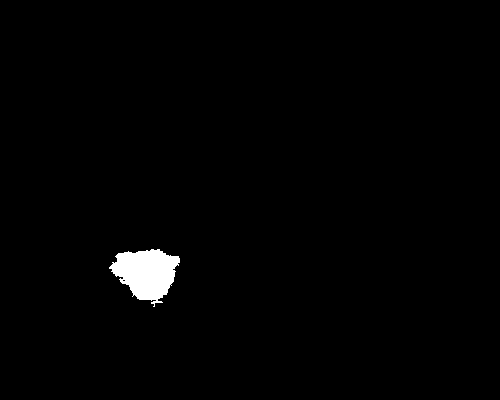} }} 
    \subfloat[]{{\includegraphics[width=2.5cm]{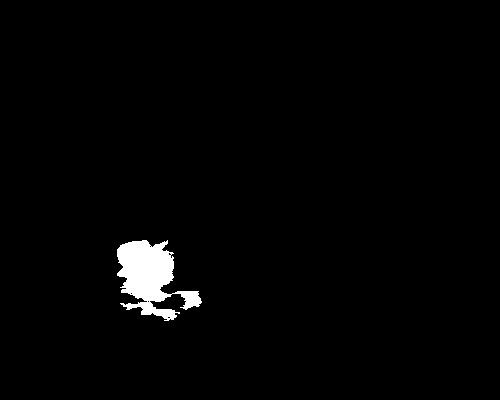} }}
    \subfloat[]{{\includegraphics[width=2.5cm]{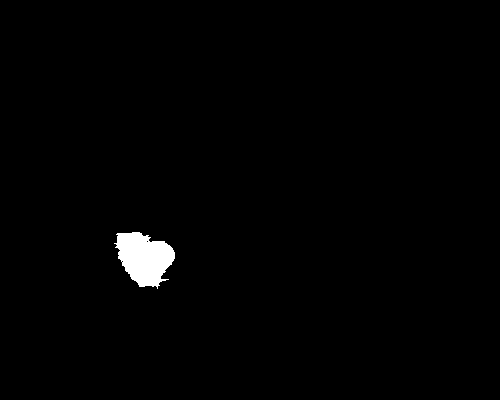} }}
    
    \subfloat[]{{\includegraphics[width=2.5cm]{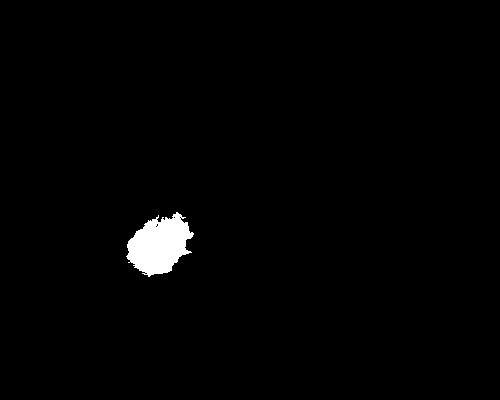} }}    
    \subfloat[]{{\includegraphics[width=2.5cm]{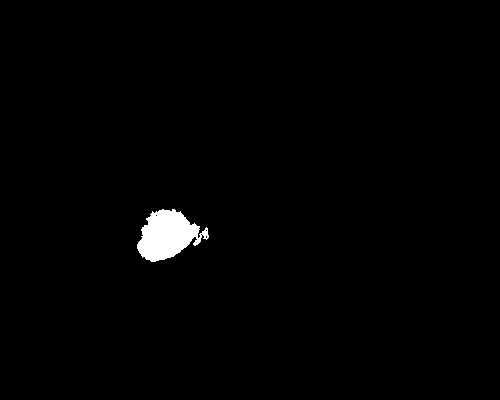} }}
    \subfloat[]{{\includegraphics[width=2.5cm]{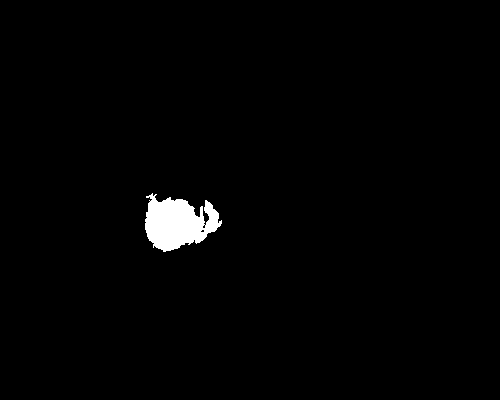} }}
    
    \subfloat[]{{\includegraphics[width=2.5cm]{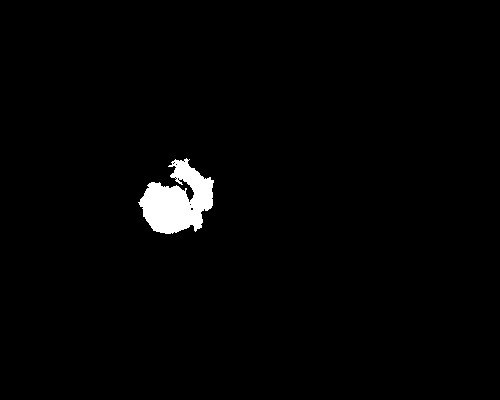} }}    
    \subfloat[]{{\includegraphics[width=2.5cm]{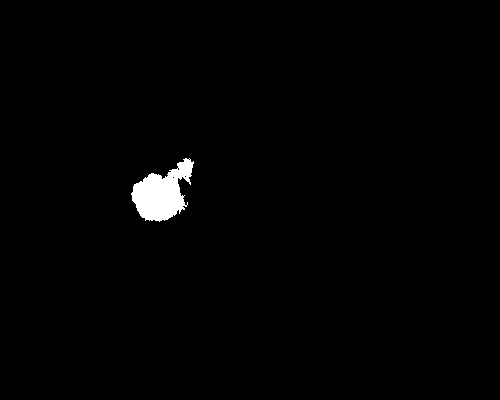} }}
    \subfloat[]{{\includegraphics[width=2.5cm]{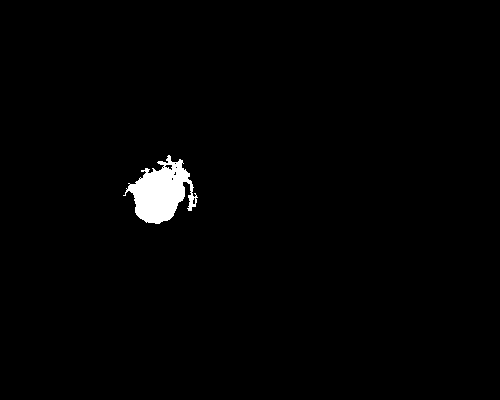} }}  

    \subfloat[]{{\includegraphics[width=2.5cm]{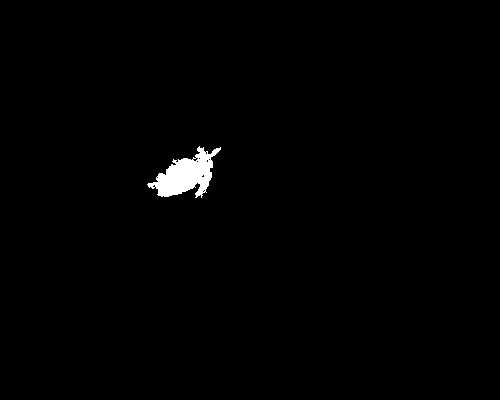} }}    
    \subfloat[]{{\includegraphics[width=2.5cm]{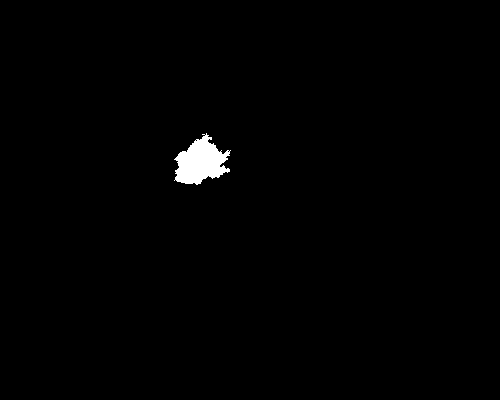} }}
    \subfloat[]{{\includegraphics[width=2.5cm]{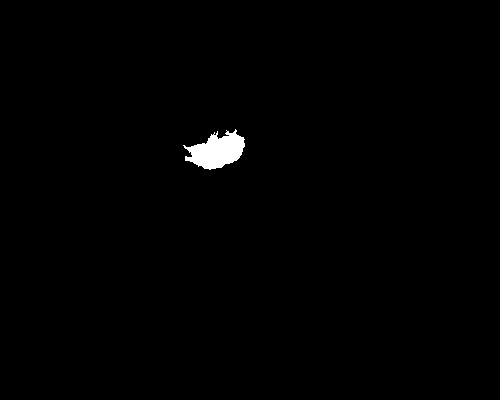} }}  
    
    \subfloat[]{{\includegraphics[width=2.5cm]{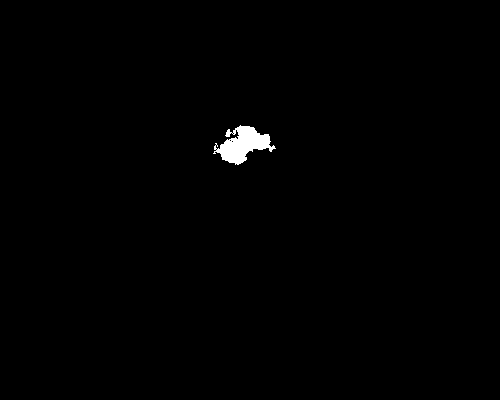} }}
    \subfloat[]{{\includegraphics[width=2.5cm]{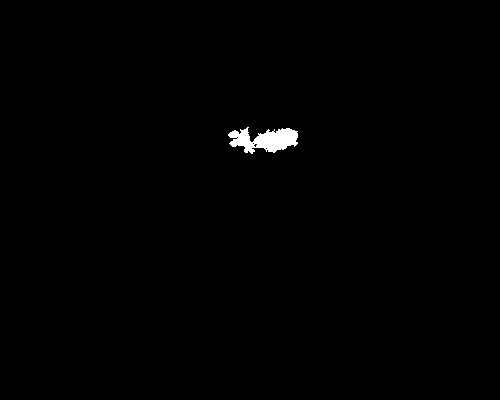} }}   
    \caption{Extracted Cyclonic Precipitation Clusters for the D\_ID (a)-(n): D1-D14}
    \label{fig6}
\end{figure}

\begin{figure}
    \centering
    \subfloat[]{{\includegraphics[width=2cm]{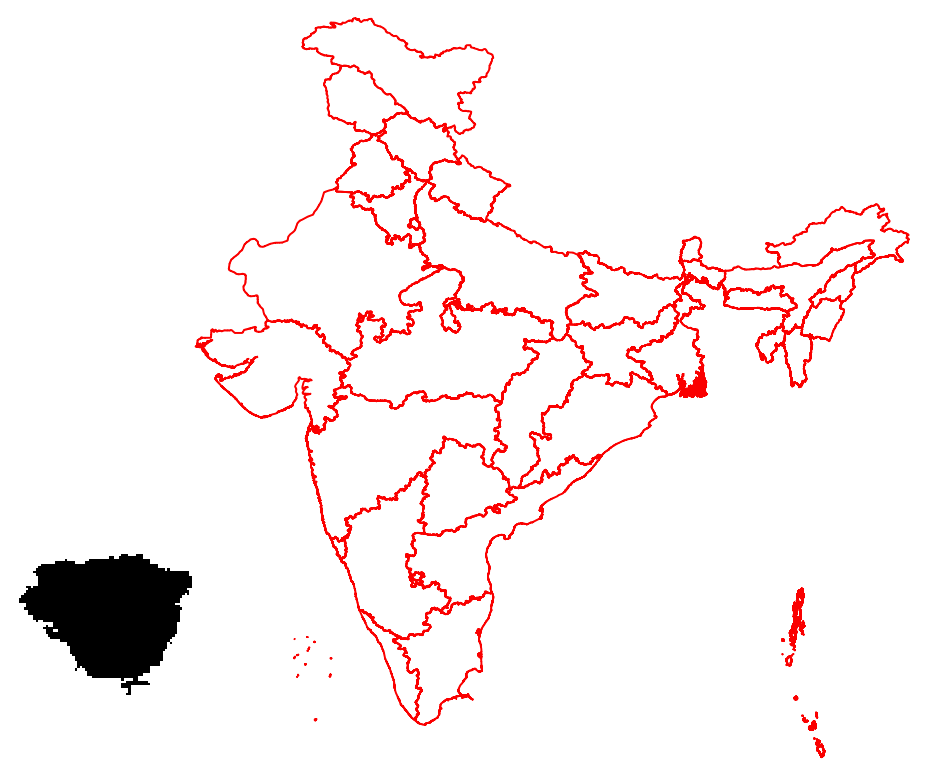} }} 
    \subfloat[]{{\includegraphics[width=2cm]{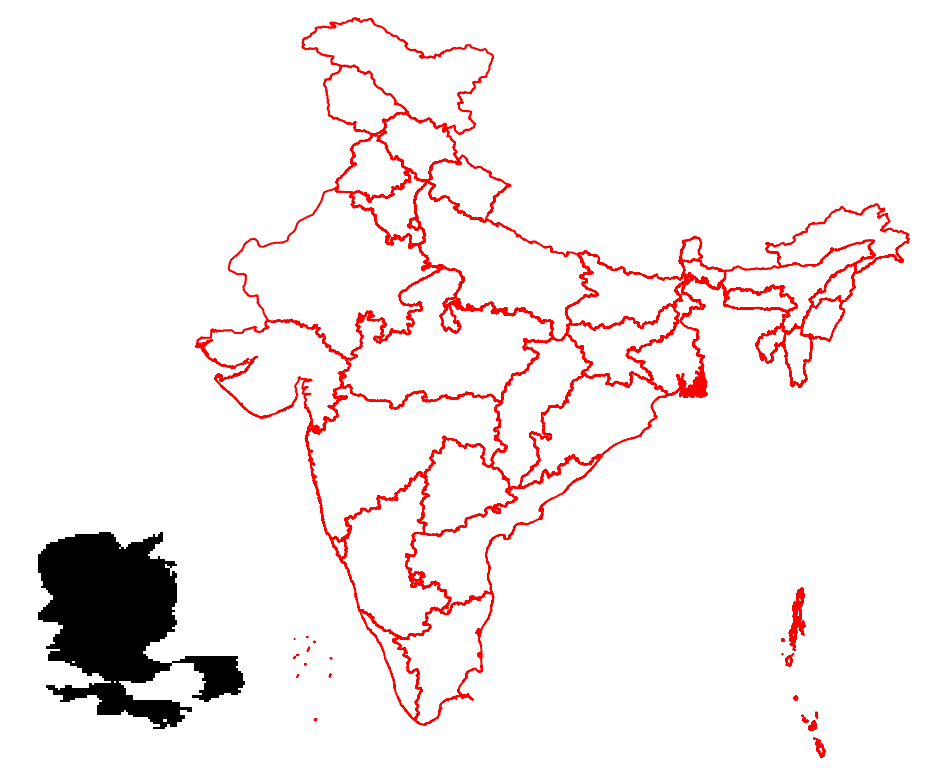} }}
    \subfloat[]{{\includegraphics[width=2cm]{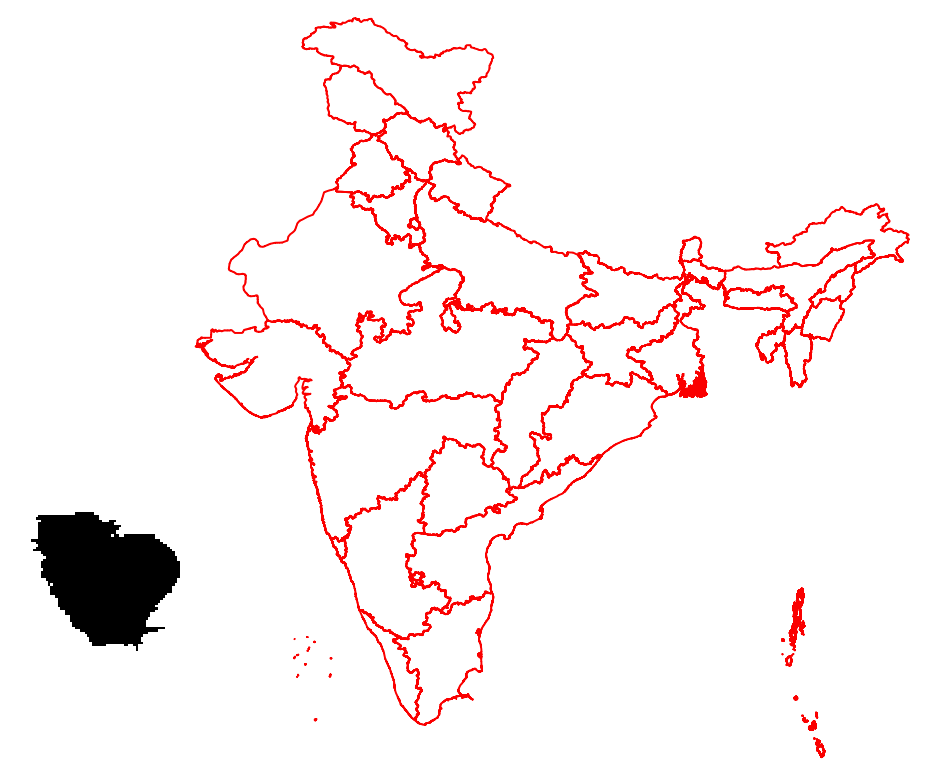} }}
    
    \subfloat[]{{\includegraphics[width=2cm]{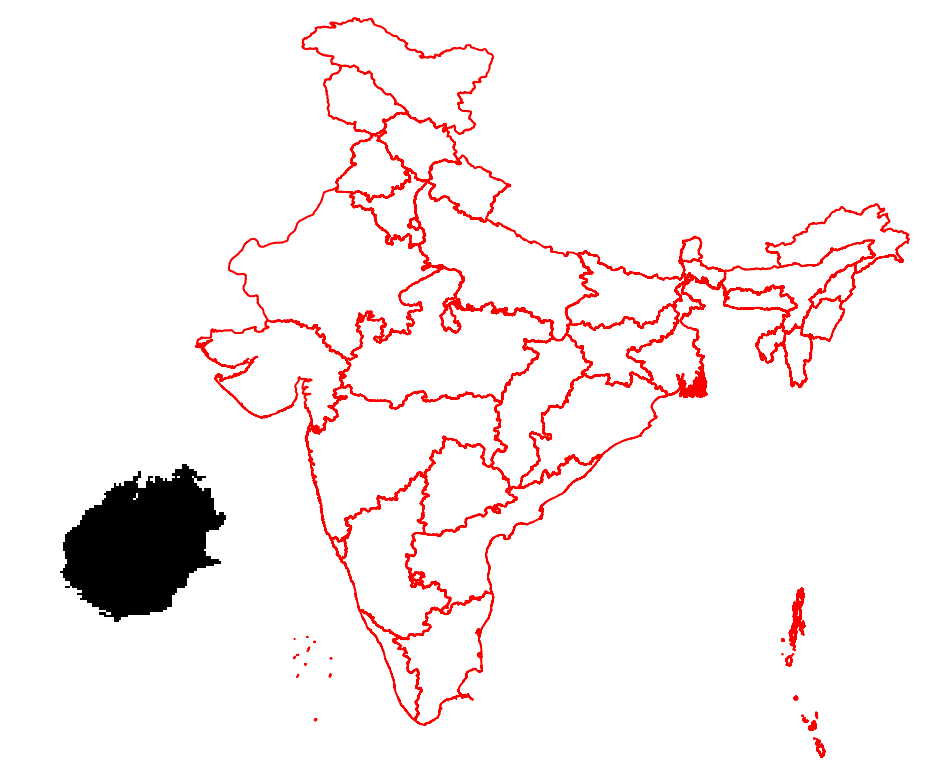} }}    
    \subfloat[]{{\includegraphics[width=2cm]{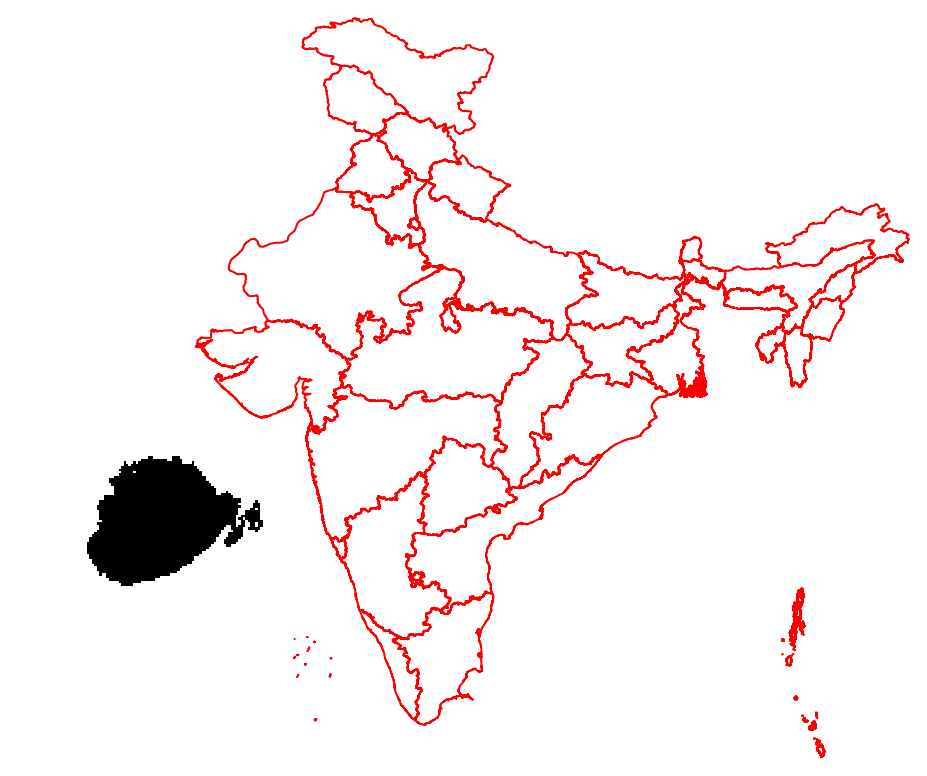} }}
    \subfloat[]{{\includegraphics[width=2cm]{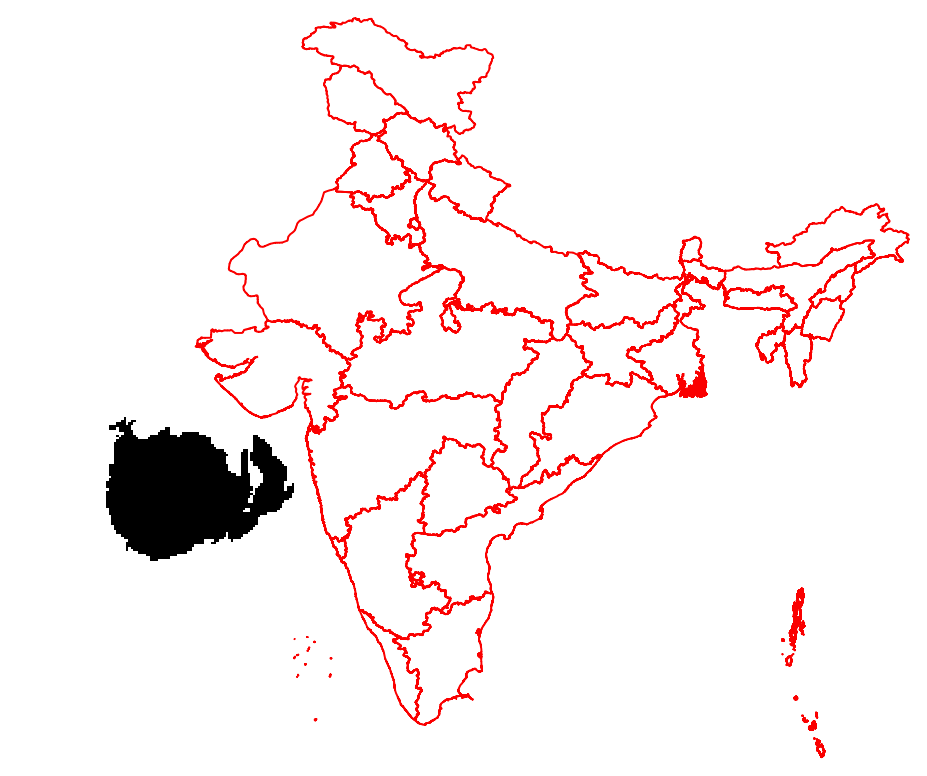} }}
    
    \subfloat[]{{\includegraphics[width=2cm]{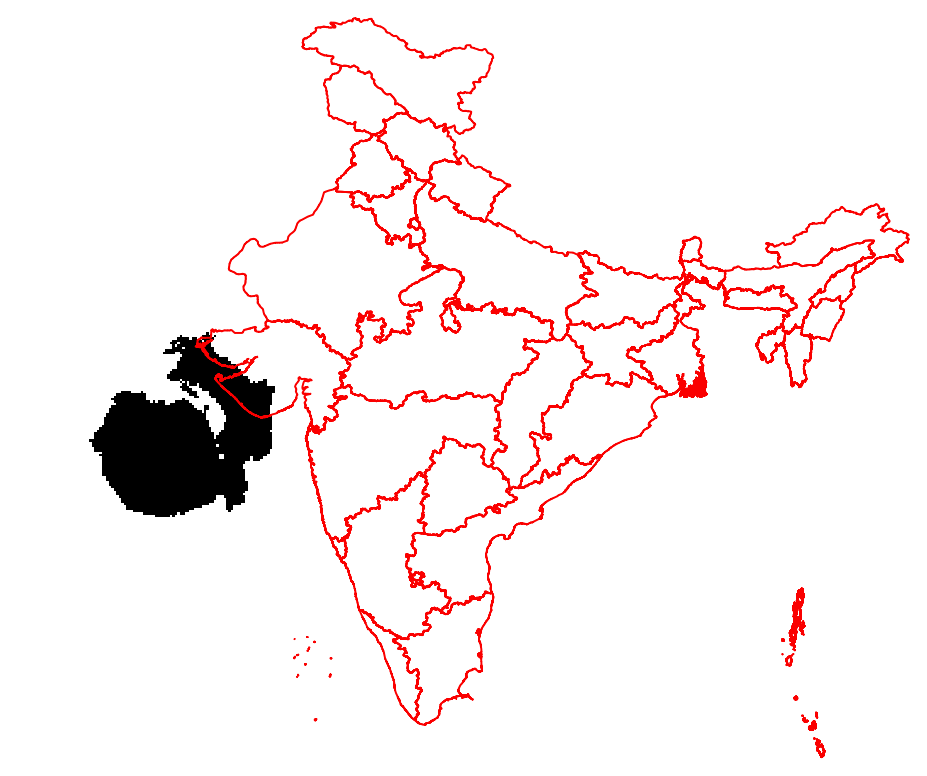} }}    
    \subfloat[]{{\includegraphics[width=2cm]{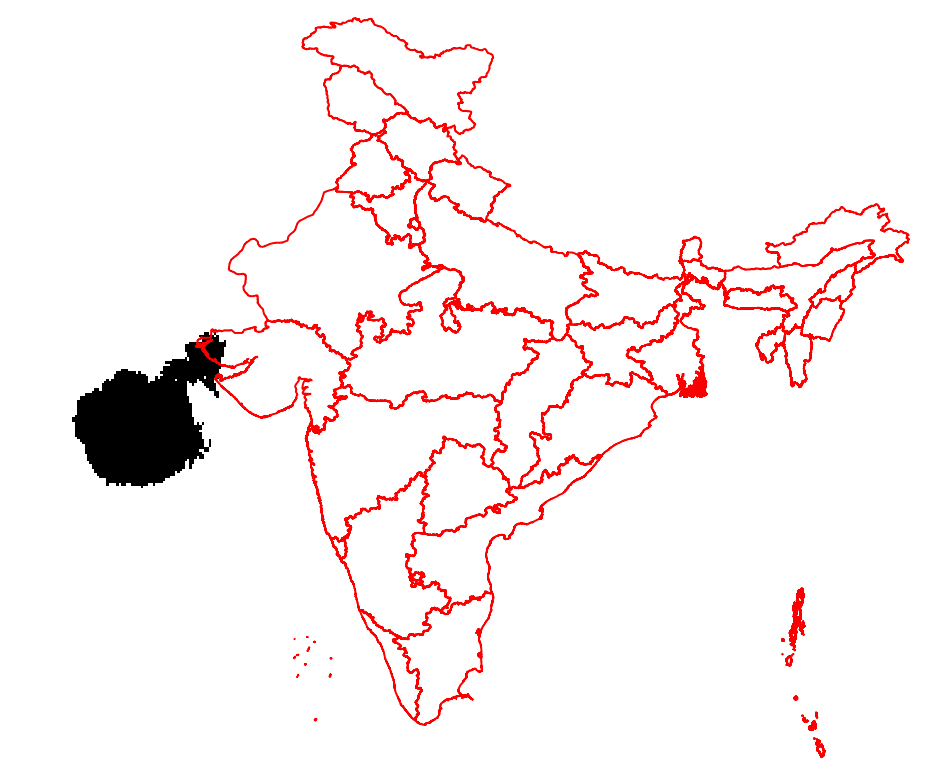} }}
    \subfloat[]{{\includegraphics[width=2cm]{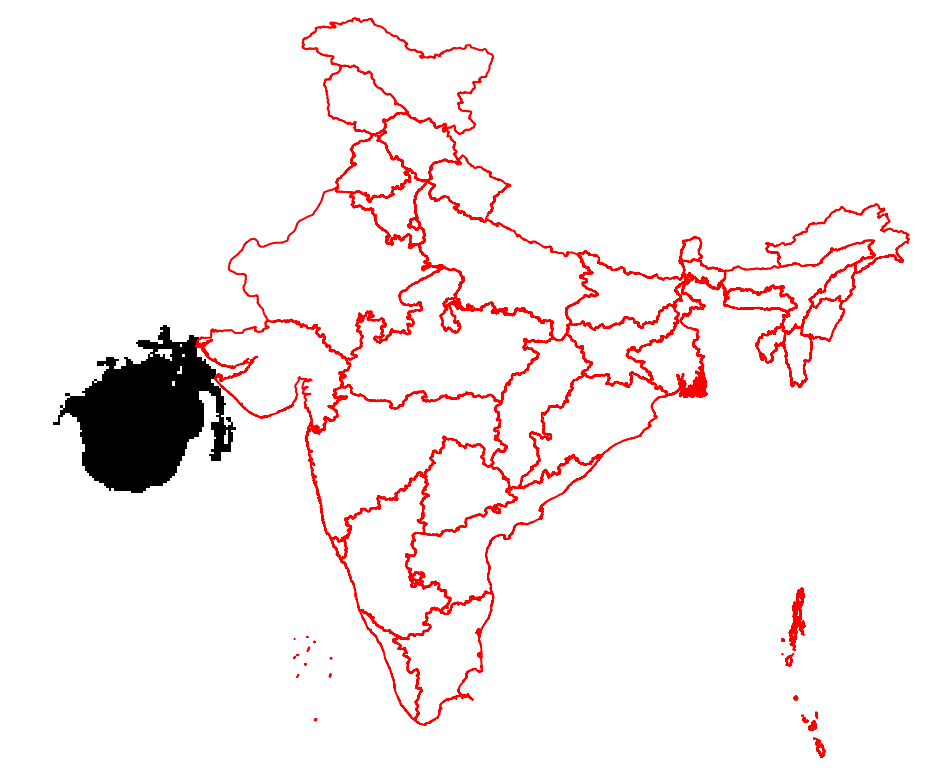} }}  

    \subfloat[]{{\includegraphics[width=2cm]{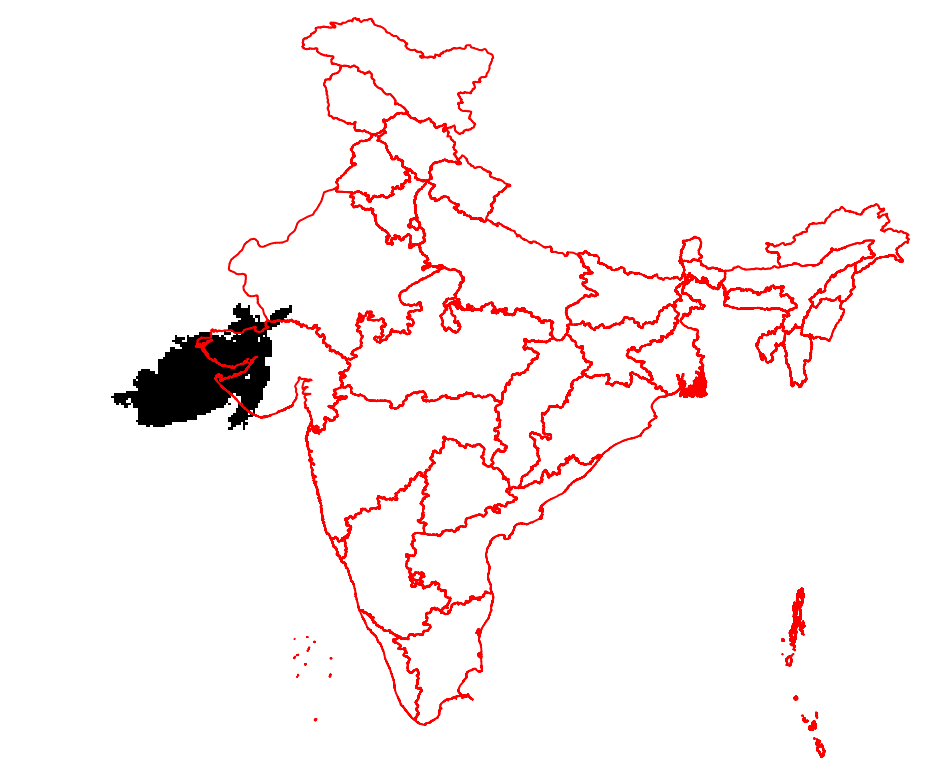} }}    
    \subfloat[]{{\includegraphics[width=2cm]{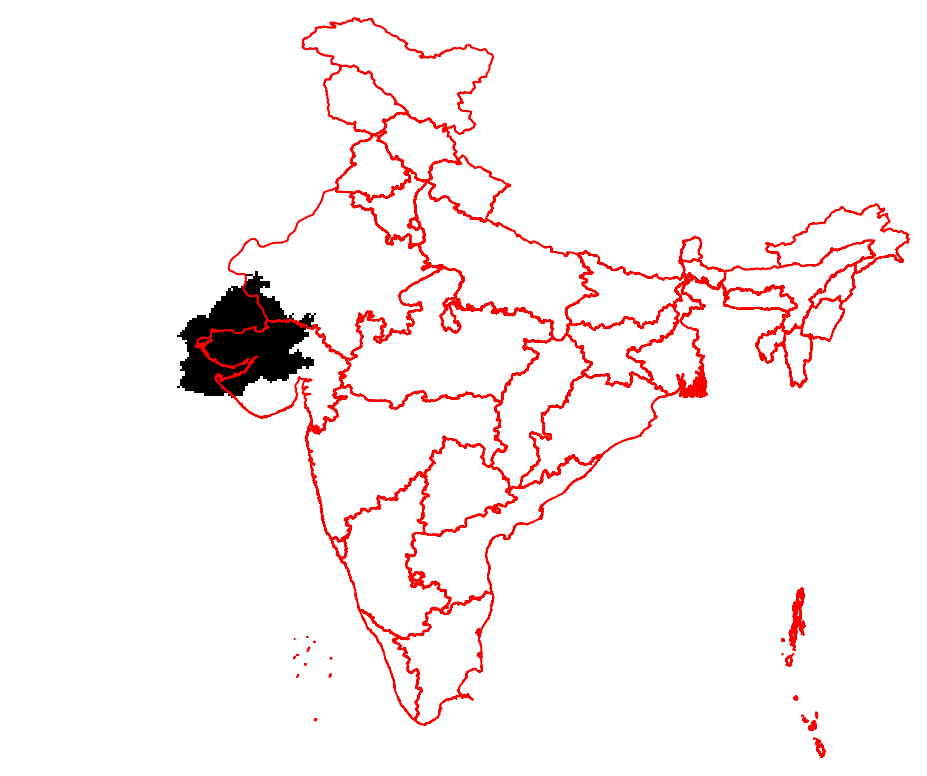} }}
    \subfloat[]{{\includegraphics[width=2cm]{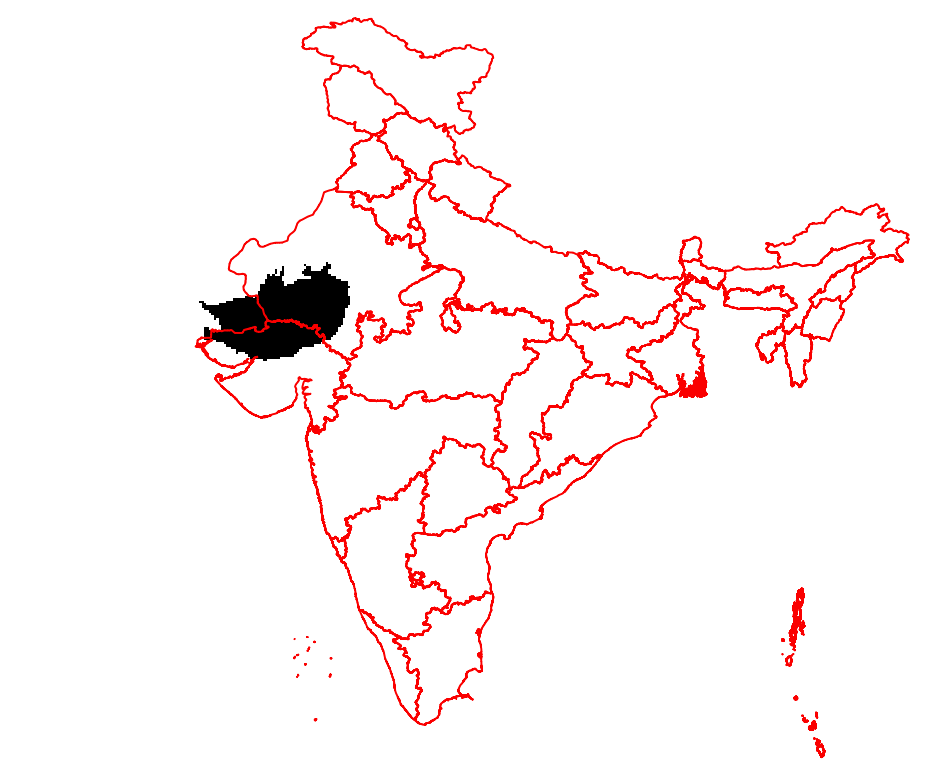} }}  
    
    \subfloat[]{{\includegraphics[width=2cm]{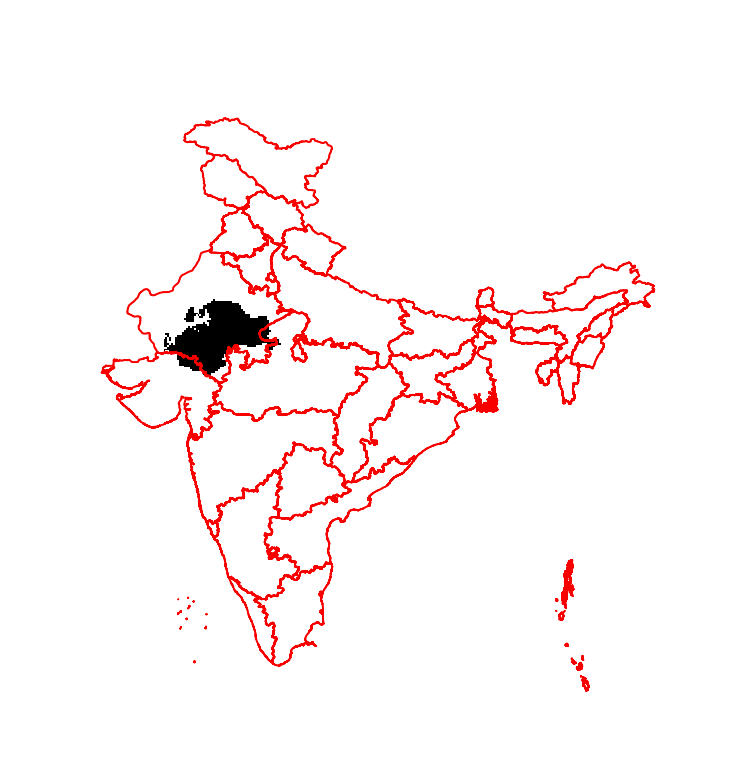} }}
    \subfloat[]{{\includegraphics[width=2cm]{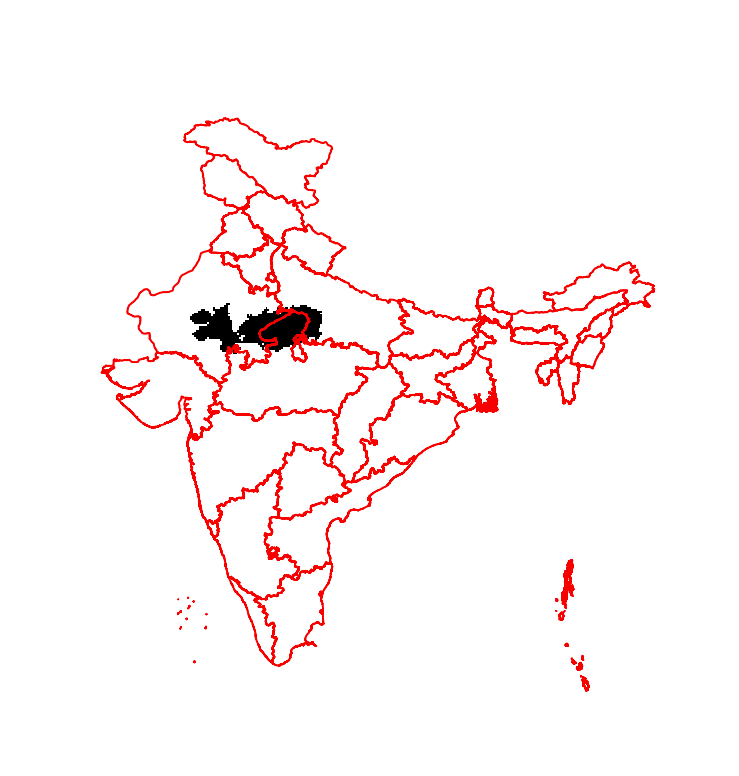} }}  
    \caption{Visualization of Extracted Cyclonic precipitation cluster over the boundary of India for the D\_ID (a)-(n): D1-D14}
    \label{fig7}
\end{figure}

The extracted images of cyclonic precipitation clusters so obtained are shown in Figure \hyperref[fig6]{\ref*{fig6}}. These images serve as pivotal visual aids for further analysis and interpretation of the cyclonic activity attributed to the Biparjoy. Moving forward, Figure \hyperref[fig7]{\ref*{fig7}}, presents a vector image of the study area projected onto the extracted cyclonic precipitation clusters, which effectively illustrates the distribution of precipitation in various Indian states.

Having identified and labeled the cyclonic precipitation clusters, the subsequent step involves computing pertinent statistics associated with these clusters. Specifically, the total precipitation associated with the Biparjoy cyclone is computed by summing up the precipitation values within the identified clusters. Additionally, the total area affected by the clusters and the total area of the affected states is also computed for each day. These computed metrics provide valuable insights into the cyclone's impact on the affected regions, offering essential information for disaster response and management efforts.

\section{Result} \label{result}

In this section, we estimate the total rainfall associated with cyclone Biparjoy on each day along with the area covered by it. Moreover, the centroid of extracted precipitation cluster and Maximum Rainfall Point are computed. Thereafter, the result is analyzed and discussed in detail. 

\subsection{Pixel Count}	
\begin{figure}[ht]
	\centering
		\includegraphics[width=8cm]{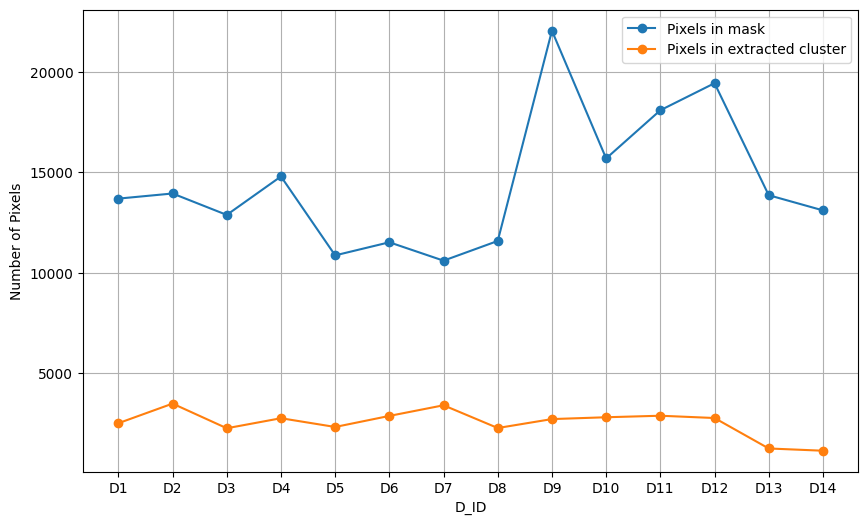}
	  \caption{Pixel Count of Mask and Extracted Cyclonic Precipitation Cluster}
        \label{fig8}
\end{figure}

The pixel counts in the processing of \ac{IMERG} product containing the number of pixels in the mask and the number of pixels in extracted precipitation clusters after applying image processing techniques from D1 to D14 are shown in Figure \hyperref[fig8]{\ref*{fig8}}.

The analysis of the pixel data in conjunction with the distinctive characteristics of Cyclone Biparjoy reveals intricate relationships between meteorological phenomena and the observed variations in precipitation clusters. On analyzing \hyperref[fig8]{\ref*{fig8}}, the findings reveal significant insights into the spatial distribution of precipitation clusters associated with the Biparjoy cyclone. Notably, the count of pixels within the binary mask consistently exceeds the count of pixels in the extracted precipitation cluster. This discrepancy is indicative of areas within the mask that do not contribute to the identified precipitation clusters. Further investigation indicates that a subset of pixels in the mask is associated with convection patterns related to the cross Equatorial southwest monsoon flow over the southeast Arabian Sea. Additionally, the rainfall observed over the Kerala and Lakshadweep area is attributed to the onset of the southwest monsoon, which commenced over most parts of Kerala and the entire Lakshadweep on June 8th.

The temporal evolution of the pixel counts provides valuable information about the dynamic nature of precipitation clusters throughout the observed period. On D9, we observe a notable increase in the number of pixels within the binary mask, suggesting an intensification of weather patterns. This escalation is likely attributed to a weather phenomenon occurring over North-Eastern part of India. Concurrently, the count of pixels in the extracted precipitation cluster remains relatively consistent, signifying a concentration of precipitation in specific regions. Furthermore, the diurnal fluctuations in convective clouds, further emphasize the dynamic nature of the precipitation clusters, reflected in the pixel counts.

Following landfall, the sustained pixel counts within the binary mask and extracted precipitation cluster illustrate the cyclone's persistence even after making landfall. These pixel trends reflect the post-landfall intensity and gradual weakening till midnight of June 16th, providing a visual representation of the cyclone's behavior.

Consistently, the pixel data reveals variations, underscoring the complexity of Cyclone Biparjoy and the challenges in predicting its path. The interconnectedness of pixel variations and meteorological characteristics highlights the potential for advanced methodologies in pixel segmentation and the integration of additional data sources for a more comprehensive understanding of cyclonic events. This holistic approach contributes to improved forecasting and enhanced disaster management capabilities.

\subsection{Estimation of Rainfall}	
Since the \ac{IMERG} product acquires data from sparse locations, the rainfall data of cyclones over the ocean is available. After pre-processing, the total rain due to the Biparjoy cyclone has been computed. Knowing the total rainfall over India after landfall is important to prepare the rainfall climatology. Therefore, the day-wise total rainfall and the corresponding area are computed before and after landfall over India by using Simple Mean Method. Though the Simple Mean method is easy to use and provides a quick estimate of the average value, it has limitations in situations where the data is highly skewed or contains outliers.

\begin{figure}
	\centering
		\includegraphics[width=80mm]{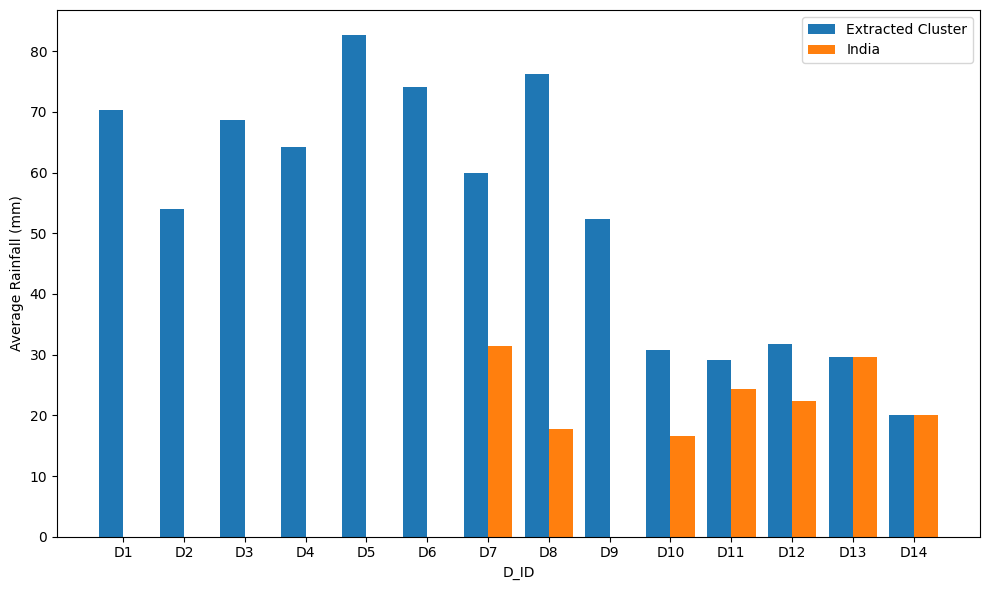}
	  \caption{Average Daily Rainfall due to Biparjoy Cyclone over India}
        \label{fig9}
\end{figure}

On the basis of extracted cyclonic precipitation cluster, the average daily rainfall is calculated over India by using Simple Mean.  In this calculation, rainfall measurements greater than 0.1mm are considered significant, while measurements less than or equal to 0.1mm are considered as trace. The result is shown in Figure \hyperref[fig9]{\ref*{fig9}}.

\begin{figure}
    \centering
    \includegraphics[width=80mm]{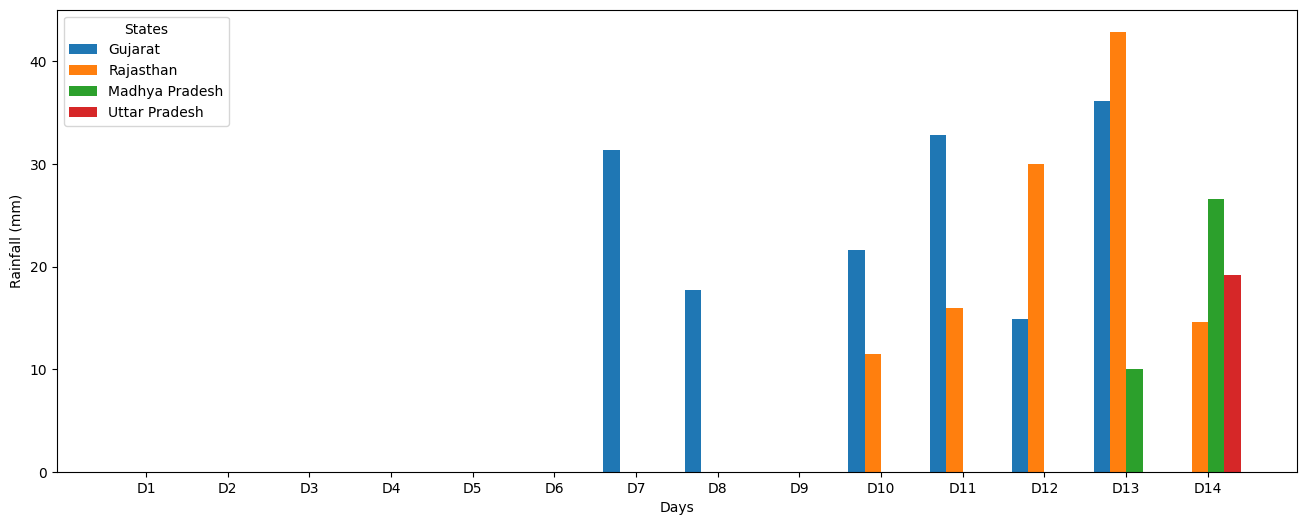}
    \caption{Rainfall received by different states of India}
    \label{fig10}
\end{figure}

Figure \hyperref[fig9]{\ref*{fig9}} provides an insight into the variation of average daily rainfall during the Biparjoy cyclone event and highlights the importance of considering both the extracted cyclonic precipitation cluster and the area over which the rainfall is averaged when analyzing rainfall patterns. 

The daily average rainfall due to Cyclone Biparjoy, computed from the extracted precipitation clusters, stands at 53.14 mm/day, encompassing both India and the Arabian Sea region, whereas the Indian boundary received an average of 11.59 mm/day. This aggregated metric serves as a key indicator of the overall intensity and extent of precipitation influenced by Cyclone Biparjoy during the analyzed timeframe. Upon closer examination of the broader geographical scope, the impact on India as a whole is characterized by significant variations in rainfall patterns across different states. 

During the influence of Cyclone Biparjoy, four states were significantly affected namely Gujarat, Rajasthan, Madhya Pradesh, and Uttar Pradesh experiencing distinct patterns of precipitation as depicted in Figure \hyperref[fig10]{\ref*{fig10}}. 
Cyclone Biparjoy made landfall on D6 along the coast of Gujarat, marking a significant meteorological event. The impact of the cyclone on Gujarat was the most pronounced impact across 6 days, marked by a substantial 31.25 mm/day downpour on D7 and an average daily rainfall of 11.03 mm/day. This rainfall persisted in Gujarat on the subsequent day, showcasing the sustained influence of Cyclone Biparjoy on the region. Intriguingly, on D9, a temporary respite from rainfall was observed over the entire Indian region. However, this reprieve was short-lived as the cyclone resumed affecting Gujarat and extended its impact to Rajasthan from the following day onwards. The meteorological data indicates a noteworthy shift in the cyclone's trajectory and influence. The absence of recorded rainfall over India on D9 signifies a brief interlude in the cyclone's activity, offering a temporary relief from precipitation across the country. Nevertheless, the subsequent resurgence of rainfall, specifically impacting Gujarat and Rajasthan, underscores the dynamic nature of Cyclone Biparjoy's trajectory and its ability to affect diverse regions over time. Rajasthan, spanning 5 days, experienced peak precipitation on D13, recording a notable 42.89 mm/day rainfall, with an average daily rainfall of 8.21 mm. Madhya Pradesh, impacted over 2 days, observed its highest precipitation of 10.05 mm on D13, featuring an average daily rainfall of 2.62 mm/day. Similarly, Uttar Pradesh, affected for 1 day, witnessed peak rainfall of 19.15 mm/day on D14, displaying a comparatively lower average daily rainfall of 1.37 mm/day. Notably, Rajasthan received the highest single-day rainfall among the four states, recording 42.89 mm/day on D13. 
This information emphasizes the localized intensity and variability of precipitation during the cyclone's impact on different regions. The state-specific statistics provide a comprehensive and detailed assessment of Biparjoy's influence on regional precipitation patterns.  These detailed insights play a pivotal role in formulating well-informed disaster management and response strategies tailored to the specific impact experienced by each state, enhancing overall preparedness and resilience.

\subsection{Estimation of Area Covered}
\begin{figure}[ht]
	\centering
		\includegraphics[width=80mm]{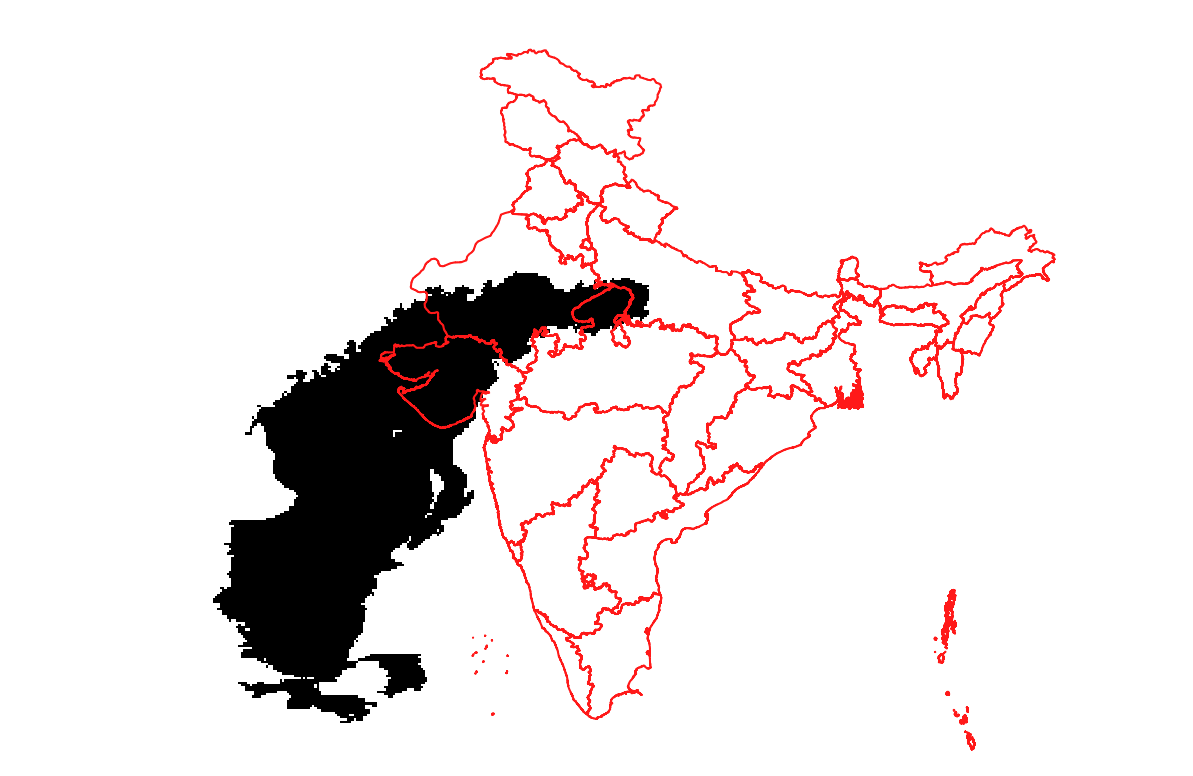}
	  \caption{Area Covered by Biparjoy Cyclone over Arabian Sea and India}\label{fig11}
\end{figure}

In this step, we conducted a thorough analysis of the spatial distribution of rainfall during the Biparjoy cyclone, aiming to understand its impact on different regions. To achieve this, we calculated the total area covered by the precipitation cluster over India. The results are visually presented in Figure \hyperref[fig11]{\ref*{fig11}}, providing valuable insights into the geographical extent of rainfall associated with the cyclone.

The comprehensive area covered by the Biparjoy cyclone was determined by unifying all the extracted raster images. The union operation allowed us to merge these individual clusters, providing a consolidated view of the spatial extent of precipitation associated with the cyclone. This approach ensures that the cumulative impact area is accurately captured, considering the evolving nature of the cyclone over time. This methodology contributes to a comprehensive analysis of the cyclone's influence on various regions, laying the groundwork for a thorough understanding of its spatial and temporal characteristics.

\begin{figure}[ht]
	\centering
		\includegraphics[width=80mm]{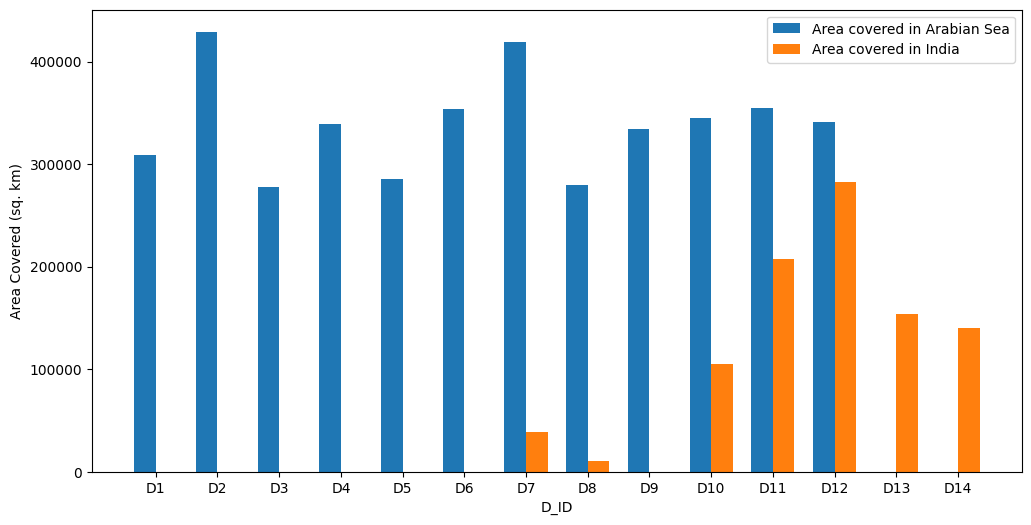}
	  \caption{Total Area Covered by Extracted Precipitation Cluster of Biparjoy Cyclone and Over India}\label{fig12}
\end{figure}

In the course of our analysis of Cyclone Biparjoy's impact on different regions, we meticulously examined the daily coverage area across Gujarat, Rajasthan, Madhya Pradesh, and Uttar Pradesh. The visual representation is shown in \hyperref[fig11]{\ref*{fig12}}.

Our findings reveal that the Biparjoy covered a substantial 411.76 thousand square kilometers in total, with Gujarat experiencing an impact over 154.75 thousand square kilometers, followed by Rajasthan with 194.80 thousand square kilometers. Madhya Pradesh and Uttar Pradesh were also significantly affected, with impact areas of 40.78 and 21.43 thousand square kilometers, respectively. 

Throughout the progression of Cyclone Biparjoy, distinct patterns of precipitation coverage were observed. On D7 and D8, a notable area of Gujarat, totaling 39.1 thousand sq. km, experienced the impact of the cyclone. Subsequently, the influence intensified on D10 and D11, extending to cover substantial regions in Gujarat, Rajasthan, and Madhya Pradesh. The zenith of the cyclone's impact was observed on D11, with widespread precipitation exceeding 150.7 thousand sq. km in Gujarat and 57.3 thousand sq. km in Rajasthan. This impactful trend persisted on D12 and D13, revealing significant coverage in Gujarat, Rajasthan, and Madhya Pradesh.

Madhya Pradesh, in particular, witnessed substantial coverage from D12 to D14, recording precipitation areas of 207.0 thousand sq. km, 138.9 thousand sq. km, and 77.9 thousand sq. km, respectively. Even on the concluding day (D14), as the cyclone's influence gradually diminished, a considerable area remained covered in Rajasthan, Madhya Pradesh, and Uttar Pradesh, highlighting the lasting effect of Cyclone Biparjoy across these regions. Notably, Rajasthan sustained coverage of 40.8 thousand sq. km, while Uttar Pradesh experienced precipitation over an area of 21.4 thousand sq. km. This detailed analysis provides valuable insights into the temporal progression of the cyclone and its varying impact on different states, contributing to our understanding for further research and disaster management strategies.

\section{Discussion} \label{discussion}

In this study, we estimated the quantitative precipitation associated with cyclone Biparjoy which provided insights into the spatial distribution of the rainfall and its impact on the affected areas. We used image processing techniques on \ac{IMERG} raster data to identify and extract precipitation clusters associated with the Biparjoy cyclone. Specifically, we computed the total rainfall for each day and also determined the area covered by the cyclone. We then proceeded to conduct a detailed analysis of the results obtained. 

The findings of this study provide a significant contribution to the understanding of Cyclone Biparjoy's impact on precipitation patterns across India. The daily average rainfall of 53.14 mm/day reflects the cyclone's substantial influence, with Gujarat experiencing a particularly pronounced impact, marked by a peak daily rainfall of 31.25 mm on D7. The dynamic trajectory of the cyclone resulted in varying levels of impact on different states, showcasing the complexity of cyclonic behavior. The spatial distribution analysis further underscores the extensive coverage area of 411.76 thousand square kilometers, emphasizing the far-reaching impact of Cyclone Biparjoy.

The temporal progression of the cyclone's influence, as revealed through this study, sheds light on the resilience and adaptability of disaster management strategies. The localized intensity and variability observed in states like Gujarat, Rajasthan, Madhya Pradesh, and Uttar Pradesh highlight the need for tailored response measures to address the diverse impacts of cyclones on different regions. The short-lived reprieve from rainfall on D9 and the subsequent resurgence underscore the challenges in predicting and managing cyclonic behavior.

This study's insights provide a foundation for future research and contribute to the development of more targeted disaster preparedness and response strategies. The nuanced understanding of cyclonic dynamics, as presented in this research, is crucial for enhancing the overall resilience of communities and infrastructure in the face of such meteorological events. 

\section{Conclusion} \label{conclusion}

Our study highlighted the significant impact of tropical cyclone Biparjoy on precipitation over India and the need for further research to better understand the dynamics and implications for regional precipitation.  This study urges further investigations to improve predictive models, enhance disaster readiness, and develop specific responses. Such efforts are vital for building resilience against the diverse impacts of tropical cyclones in the changing climate scenario.

\section*{Acknowledgement}

The author would like to express my heartfelt gratitude to the India Meteorological Development (\ac{IMD}) and the Indian Institute of Information Technology, Allahabad (IIIT Allahabad) for their invaluable support and contributions to this journal. Their guidance, resources, and assistance have been instrumental in the successful completion of this research. 

\section*{Abbreviations}

\begin{acronym}
    \acro{AMSU}{Advanced Microwave Sounding Unit}
    \acro{BMTPC}{Building Materials and Technology Promotion Council}
    \acro{BoB}{Bay of Bengal}
    \acro{CCA}{Connected Component Analysis}
    \acro{CS}{Cyclonic Storm}
    \acro{D}{Depression}
    \acro{DOA}{Date of Acquisition}
    \acro{ESCS}{Extremely Severe Cyclonic Storm}
    \acro{GEO}{Geosynchronous-Earth-Orbit}
    \acro{GPM}{Global Precipitation Measurement}
    \acro{IMD}{India Meteorological Department}
    \acro{IMERG}{Integrated Multi-satellitE Retrievals for GPM}
    \acro{IR}{Infrared}
    \acro{JAXA}{Japan Aerospace Exploration Agency}
    \acro{JTWC}{Joint Typhoon Warning Centre}
    \acro{LEO}{Low-Earth-Orbit}
    \acro{MSW}{Maximum Sustained Wind}
    \acro{NASA}{National Aeronautics and Space Administration}
    \acro{NDMA}{National Disaster Management Authority}
    \acro{NIO}{North Indian Ocean}
    \acro{NWP}{Numerical Weather Prediction}
    \acro{ROI}{Region of Interest}
    \acro{TC}{Tropical Cyclones}
    \acro{TCs}{Tropical Cyclones}
\end{acronym}

\bibliographystyle{model1-num-names}
\bibliography{TC-Rainfall-Biparjoy.bib}

\end{document}